%% file: main.tex
\documentclass{article}
\usepackage[preprint, nonatbib]{neurips_2021}
\usepackage[round]{natbib}
\usepackage[utf8]{inputenc} % allow utf-8 input
\usepackage[T1]{fontenc}    % use 8-bit T1 fonts
\usepackage[colorlinks=true,linkcolor=magenta,citecolor=green!80!black]{hyperref}
\usepackage{url}            % simple URL typesetting
\usepackage{booktabs}       % professional-quality tables
\usepackage{amsfonts}       % blackboard math symbols
\usepackage{nicefrac}       % compact symbols for 1/2, etc.
\usepackage{microtype}      % microtypography
\usepackage{multicol}
\usepackage{pifont}
\usepackage{physics}
\usepackage{wrapfig}
\usepackage[toc,page,header]{appendix}
\usepackage{minitoc}
\numberwithin{equation}{section}
\usepackage{etoolbox}
\usepackage{fancyhdr}
\usepackage{stmaryrd}
\usepackage[dvipsnames]{xcolor}
\usepackage{accents}
\usepackage{newunicodechar}
\usepackage[bb=boondox,bbscaled=.95]{mathalfa}
\usepackage{amsmath, amssymb, amsfonts, amsthm, bm}
\usepackage{physics}
\usepackage{cancel}
\usepackage{thmtools, thm-restate}

\declaretheorem{proposition}

\newtheorem{definition}{Definition}

\usepackage[frozencache, cachedir=./minted]{minted} % [finalizecache, cachedir=./minted]{minted}  %
\usepackage{tcolorbox}
\tcbuselibrary{minted,breakable,xparse,skins}
\definecolor{bg}{gray}{0.95}

\DeclareTCBListing{mintedbox}{O{}m!O{}}{%
  breakable=true,
  listing engine=minted,
  listing only,
  minted language=#2,
  minted style=default,
  minted options={%
    linenos,
    gobble=0,
    breaklines=true,
    breakafter=,,
    fontsize=\small,
    numbersep=8pt,
    #1},
  boxsep=0pt,
  left skip=0pt,
  right skip=0pt,
  left=25pt,
  right=0pt,
  top=3pt,
  bottom=3pt,
  arc=5pt,
  leftrule=0pt,
  rightrule=0pt,
  bottomrule=2pt,
  toprule=2pt,
  colback=bg,
  colframe=blue!70!white,
  enhanced,
  overlay={%
    \begin{tcbclipinterior}
    \fill[blue!20!white] (frame.south west) rectangle ([xshift=20pt]frame.north west);
    \end{tcbclipinterior}},
  #3}

\newcommand{\ubar}[1]{\underaccent{\bar}{#1}}
\newcommand{\x}{\times}
\DeclareMathOperator{\argmin}{arg min}
\newcommand{\cD}{\mathcal{D}}
\newcommand{\cE}{\mathcal{E}}
\newcommand{\cN}{\mathcal{N}}

\newcommand{\cT}{\mathcal{T}}
\newcommand{\cX}{\mathcal{X}}
\newcommand{\cY}{\mathcal{Y}}
\newcommand{\cW}{\mathcal{W}}
\newcommand{\cZ}{\mathcal{Z}}
\newcommand{\sD}{\mathsf{D}}
\newcommand{\bE}{\mathbb{E}}
\newcommand{\R}{\mathbb{R}}

\newcommand{\nZ}{{n_z}}

\newcommand{\nT}{{n_\theta}}

\newcommand{\0}{\mathbb{0}}
\newcommand{\Id}{\mathbb{I}}

\def\blue{\color{blue!70!white}}
\def\ocra{\color{ocra}}
\definecolor{olive}{rgb}{0.6, 0.6, 0.2}
\definecolor{sand}{rgb}{0.8666666666666667, 0.8, 0.4666666666666667}
\definecolor{wine}{rgb}{0.5333333333333333, 0.13333333333333333, 0.3333333333333333}
\definecolor{deblue}{RGB}{11,132,147}
\definecolor{degray}{RGB}{150,150,156}
\definecolor{ocra}{RGB}{204, 119, 34}

\newcommand{\fsquare}[2][red,fill=red]{\tikz[baseline=-0.5ex]\draw[#1,radius=#2] (-0.1,-0.1) rectangle (0.1,0.1);}

\newcommand*{\colorboxed}{}
\def\colorboxed#1#{%
  \colorboxedAux{#1}%
}
\newcommand*{\colorboxedAux}[3]{%
  \begingroup
    \colorlet{cb@saved}{.}%
    \color#1{#2}%
    \boxed{%
      \color{cb@saved}%
      #3%
    }%
  \endgroup
}
\newcommand{\inner}[3][]{\ensuremath{\left\langle #2, \, #3 \right\rangle_{#1}}}
\newunicodechar{λ}{{$\mathtt\lambda$}}
\newunicodechar{μ}{{$\mathtt\mu$}}

\title{Differentiable Multiple Shooting Layers}
\author{
Stefano Massaroli$^{1,}$\thanks{Equal contribution. Author order was decided by flipping a coin. $^1$The University of Tokyo, $^2$KAIST. $^3$RIKEN. Corresponding author: \textit{Stefano Massaroli}, email: {$\tt massaroli@robot.t.u-tokyo.ac.jp$}} , Michael Poli$^{2,*}$,\\ \textbf{Sho Sonoda$^{3}$, Taiji Suzuki$^{1,3}$, Jinkyoo Park$^{2}$, Atsushi Yamashita$^{1}$, Hajime Asama$^{1}$}
}
\begin{document}
\maketitle
\doparttoc
\faketableofcontents

\vspace{-2mm}
\begin{abstract}
    \vspace{-2mm}
    We detail a novel class of implicit neural models. Leveraging time--parallel methods for differential equations, \textit{Multiple Shooting Layers} (MSLs) seek solutions of initial value problems via parallelizable root-finding algorithms. MSLs broadly serve as drop--in replacements for \textit{neural ordinary differential equations} (Neural ODEs) with improved efficiency in number of function evaluations (NFEs) and wall--clock inference time. We develop the algorithmic framework of MSLs, analyzing the different choices of solution methods from a theoretical and computational perspective. MSLs are showcased in long horizon optimal control of ODEs and PDEs and as latent models for sequence generation. Finally, we investigate the speedups obtained through application of MSL inference in neural controlled differential equations (Neural CDEs) for time series classification of medical data.
\end{abstract}
\input{Chapters/1_Introduction}

\input{Chapters/2_Mshooting_Layers}
\input{Chapters/3_Realization}
\input{Chapters/4_Applications}
\input{Chapters/5_Discussion}

%
%%%%%%%%%%%%%%
%Bibliography
\bibliographystyle{abbrvnat}
\bibliography{main.bib}
%
%%%%%%%%%%%%%%%%%%%%%%%%%
% APPENDIX
%%%%%%%%%%%%%%%%%%%%%%%%
%
\newpage
\rule[0pt]{\columnwidth}{3pt}
\begin{center}
    \huge{\bf{Differentiable Multiple Shooting Layers} \\
    \emph{Supplementary Material}}
\end{center}
\vspace*{3mm}
\rule[0pt]{\columnwidth}{1pt}
\vspace*{-.5in}
\appendix
\addcontentsline{toc}{section}{}
\part{}
\parttoc
\input{Appendix/Appendix_A}
\input{Appendix/Appendix_B}
\input{Appendix/Appendix_C}
\input{Appendix/Appendix_D}
\input{Appendix/Appendix_E}

\end{document}

%% file: Chapters/1_Introduction.tex
\section{Introduction}
\begin{wrapfigure}[11]{r}{0.5\textwidth}
\vspace{-6mm}
    \begin{center}
        \includegraphics[width=\linewidth]{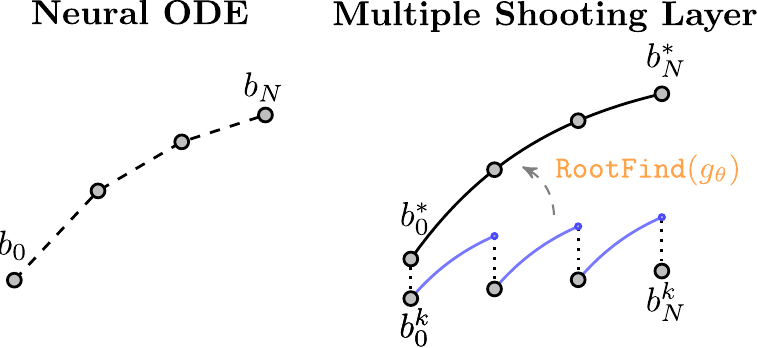}
    \end{center}
    \vspace{-4mm}
    \caption{\footnotesize MSLs apply parallelizable root finding methods to obtain differential equation solutions.}
    \label{fig:msltoy}
\end{wrapfigure}

\begin{center}\small
    \textit{For the last twenty  years, one has tried to speed up numerical computation mainly by providing ever faster computers. Today, as it appears that one is getting closer to the maximal speed of electronic components, emphasis is put on allowing operations to be performed in parallel. In the near future, much of numerical analysis will have to be recast in a more ``parallel'' form. \citealp{nievergelt1964parallel}}
\end{center}
Discovering and exploiting parallelization opportunities has allowed deep learning methods to succeed across application areas, reducing iteration times for architecture search and allowing scaling to larger data sizes \citep{krizhevsky2012imagenet,diamos2016persistent,vaswani2017attention}.
Inspired by multiple shooting, time--parallel methods for ODEs \citep{bock1984multiple,diehl2006fast,gander201550,staff2005stability} and recent advances on the intersection of differential equations, implicit problems and deep learning, we present a novel class of neural models designed to maximize parallelization across \textit{time}: differentiable \textit{Multiple Shooting Layers} (MSLs). MSLs seek solutions of \textit{initial value problems} (IVPs) as roots of a function designed to ensure satisfaction of boundary contraints. Figure \ref{fig:msltoy} provides visual intuition of the parallel nature of MSL inference. 

\paragraph{\fsquare[fill=deblue]{3pt}~An implicit neural differential equation} MSL inference is built on the interplay of numerical methods for root finding problems and differential equations. This property reveals the proposed method as a missing link between implicit--depth architectures such as Deep Equilibrium Newtorks (DEQs) \citep{bai2019deep} and continuous--depth models \citep{weinan2017proposal,NEURIPS2018_69386f6b,massaroli2020dissecting,kidger2020neural,li2020scalable}.
Indeed, MSLs can be broadly applied as drop--in replacements for Neural ODEs, with the advantage of often requiring a smaller \textit{number of function evaluations} (NFEs) for neural networks parametrizing the vector field. MSL variants and their computational signature are taxonomized on the basis of the particular solution algorithm employed, such as Newton and \textit{parareal} \citep{maday2002parareal} methods. %
\vspace*{-2mm}
\paragraph{\fsquare[fill=gray!70!orange]{3pt}~Faster inference and fixed point tracking}
Differently from classical multiple shooting methods, MSLs operate in regimes where function evaluations of the vector field can be significantly more expensive than surrounding operations. For this reason, the reduction in NFEs obtained through time--parallelization leads to significant inference speedups. In full-batch training regimes, MSLs provably enable tracking of fixed points across training iterations, leading to drastic acceleration of forward passes (often the cost of a single root finding step). We apply the tracking technique to optimal control of ODEs and PDEs, with speedups in the order of several times over Neural ODEs. MSLs are further evaluated in sequence generation via a latent variant, and as a faster alternative to \textit{neural controlled differential equations} (Neural CDE) \citep{kidger2020neural} in long--horizon time series classification.

%% file: Chapters/2_Mshooting_Layers.tex
\section{Multiple Shooting Layers}
Consider the \textit{initial value problem} (IVP)
\begin{equation}\label{eq:1}
    \begin{aligned}
    	\dot z(t) &= f_\theta(t, z(t))\\
    	z(0) &= z_0
    \end{aligned}, ~~~t\in[0, T].
\end{equation}
with state $z\in\cZ\subset\R^\nZ$, parameters $\theta\in\cW$ for some space $\cW$ of functions $[0,T]\rightarrow\R^\nT$ and a smooth vector field $f_\theta:[0,T]\times\cZ\times\cW\rightarrow\cZ$. For all $z\in\cZ,~s,t\in[0,T];s<t$ we denote with $\phi_\theta(z, s, t)$ the solution of \eqref{eq:1} at time $t$ starting from $z$ at time $s$ , i.e. $\phi_\theta(z, s, t): (z, s, t)\mapsto z(t)$.

The crux behind \textit{multiple--shooting} methods for differential equations is to turn the initial value problem \eqref{eq:1} into a \textit{boundary value problem} (BVP). We split the the time interval $[0, T]$ in $N$ sub--intervals $[t_{n}, t_{n+1}]$ with $0=t_0<t_1<\cdots<t_{N}=T$ and define $N$ left boundary subproblems
\begin{equation}\label{eq:2}
	\begin{aligned}
	z_n(t_n) = {\blue b_n}~~\text{and}~~\dot z_n(t) = f_\theta(t, z_n(t))
    \end{aligned},~t\in[t_n, t_{n+1}]%,~n=0, 1, \dots, N-1.
\end{equation}
where ${\blue b_n}$ are denoted as \textit{shooting parameters}. At each time $t\in[0, T]$, the solution of $\eqref{eq:2}$ matches the one of $\eqref{eq:1}$ iff all the shooting parameters $b_n$ are identical to $z(t_n)$, $b_n = \phi_\theta(z_0, t_0, t_n)$. Using $z(t_n) = \phi_\theta(z(t_{n-1}), t_{n-1}, t_n)$, we obtain the equivalent conditions
$$
	\begin{aligned}
	b_0 &=\phi_\theta(z_0, t_0,t_0) = z_0\\
	b_1 &=\phi_\theta(b_0, t_0, t_1) = z_0(t_1)\\
	&~~\vdots\\
	b_{N} &=\phi_\theta(b_{N-1}, t_{N-1}, t_{N})= z_{N-1}(t_N)\\
	\end{aligned}
$$
Let $B := (b_0, b_1,\cdots ,b_N)$ and $\gamma_\theta(B, z_0) := \left(z_0, \phi_\theta(b_0, t_0, t_1), \cdots, \phi_\theta(b_{N-1}, t_{N-1}, t_{N})\right)$. We can thus turn the IVP \eqref{eq:1} into the roots--finding problem the of a function $g_\theta$ defined as 
$$
    g_\theta(B, z_0) = B - \gamma_\theta(B, z_0)
$$
\begin{tcolorbox}[enhanced, drop fuzzy shadow]
    \begin{definition}[Multiple Shooting Layer (MSL)] With $\ell_x:\cX\rightarrow\cZ$ and $\ell_y:\cZ^{N+1}\rightarrow\cY$ two affine maps, a {$\tt multiple~shooting~layer$} is defined as the implicit input--output mapping ${\ocra x}\mapsto {\blue y}$:
    \begin{equation}\label{eq:3}
         \begin{aligned}
            z_0 &= \ell_x({\ocra x})\\
            B^* &~:~g_\theta(B^*, z_0) = \0\\
            {\blue y} &=  \ell_y(B^*)
         \end{aligned}
    \end{equation}
    \end{definition}
\end{tcolorbox}

%% file: Chapters/3_Realization.tex
\section{Realization of Multiple Shooting Layers}
The remarkable property of MSL is the possibility of computing the solutions of all the $N$ IVPs \eqref{eq:2} in parallel from the shooting parameters in $B$ with any standard ODE solver. This allows for a drastic reduction in the number of vector field evaluations at the cost of a higher memory requirement for the parallelization to take place. Nonetheless, the forward pass of MSLs requires the shooting parameters $B$ to satisfy the nonlinear algebraic matching condition $g_\theta(B, z_0)=\0$, which has also to be solved numerically. 
\subsection{Forward Model}\label{sec:3.1}
The \textit{forward} MSL model involves the synergistic combination of two main classes of numerical methods: \textit{ODE solvers} and \textit{root finding} algorithms, to compute $\gamma_\theta(B, z_0)$ and $B^*$, respectively. There exists a hierarchy between the two classes of methods: the ODE solver will be invoked at each step $k$ of the root finding algorithm to compute $\gamma_\theta (B^k, z_0)$ and evaluate the matching condition $g_\theta(B^k, z_0)$.
\paragraph{Newton methods for root finding}
Let us denote with $B^k$ the solution of the root finding problem at the $k$-th step of the Newton method and let $\sD g_\theta (B^k, z_0)$ be the Jacobian of $g_\theta$ computed in $B^k$. The solution $B^*~:~g_\theta(B^*, z_0)=\0$ can be obtained by iterating the Newton--Raphson fixed point iteration
\begin{equation}\label{eq:5}
        B^{k+1} = B^k - \alpha \left[\Id_{N}\otimes \Id_{n_z} - \sD\gamma_\theta(B^k,z_0)\right]^{-1}\left[B^k - \gamma_\theta(B^k, z_0)\right]
\end{equation}
which converges quadratically to $B^*$ \citep{nocedal2006numerical}. The exact Newton iteration \eqref{eq:5} theoretically requires the inverse of the Jacobian $\Id_{N}\otimes \Id_{n_z} - \sD\gamma_\theta(B^k,z_0)$. Without the special structure of the MSL problem, the Jacobian would have had to be the computed in full, as in the case of DEQs \citep{bai2019deep}. Being the Jacobian of dimension $\R^{Nn_z\times Nn_z}$, its computation with \textit{reverse--mode} automatic differentiation (AD) tools scales poorly with state dimensions and number of shooting parameter (cubically in both $n_z$ and $N$).
\begin{wrapfigure}[7]{r}{0.45\textwidth}
    \vspace{-6mm}
    \begin{center}
        \includegraphics[scale=.95]{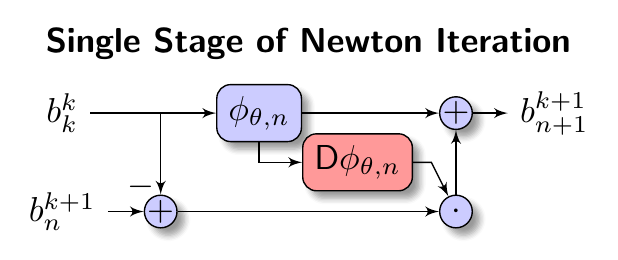}
    \end{center}
    \vspace{-6mm}
    \caption{\footnotesize One stage of Newton iteration \eqref{eq:7}.}%
    \label{fig:newt_common}
\end{wrapfigure}
Instead, the special structure of the MSL matching function $g_\theta(B, z_0) = B - \gamma_\theta(B, z_0)$ and its Jacobian, opens up application of direct updates where inversion is not required. 

\paragraph{Direct multiple shooting}

Following the treatment of \cite{chartier1993parallel}, we can obtain a \textit{direct} formulation of the Newton iteration which does not require the composition of the whole Jacobian nor its inversion. The direct multiple shooting iteration is derived by setting $\alpha=1$ and multiplying the Jacobian on both sides of \eqref{eq:5} yielding
\[
     \left[\Id_{N}\otimes \Id_{n_z} - \sD\gamma_\theta(B^k,z_0)\right](B^{k+1} - B^{k}) = \gamma_\theta(B^k, z_0) - B^k
\]
which leads to the following update rule for the individual shooting parameters $b_n^k$ (see Fig. \ref{fig:newt_common}):
\begin{equation}\label{eq:7}
        \begin{aligned}
        b_{n+1}^{k+1} = \phi_{\theta,n}(b_n^k) + \sD\phi_{\theta, n}(b_n^k)\left({\blue b_n^{k+1}} - b_n^k\right),~~b_0^{k+1} = z_0
    \end{aligned}
\end{equation}
where $\sD\phi_{\theta, n}(b_n^k) = \dd\phi_{\theta, n}(b_n^k)/\dd b_n $ is the sensitivity of each individual flow to its initial condition. Due to the dependence of $b_{n+1}^{k+1}$ on $\blue b_n^{k+1}$, a complete Newton iteration theoretically requires $N-1$ sequential stages.

\begin{wrapfigure}[20]{l}{0.51\textwidth}
    \vspace{-8.5mm}
    \begin{center}
        \includegraphics[scale=.9, trim={10 0 0 0}]{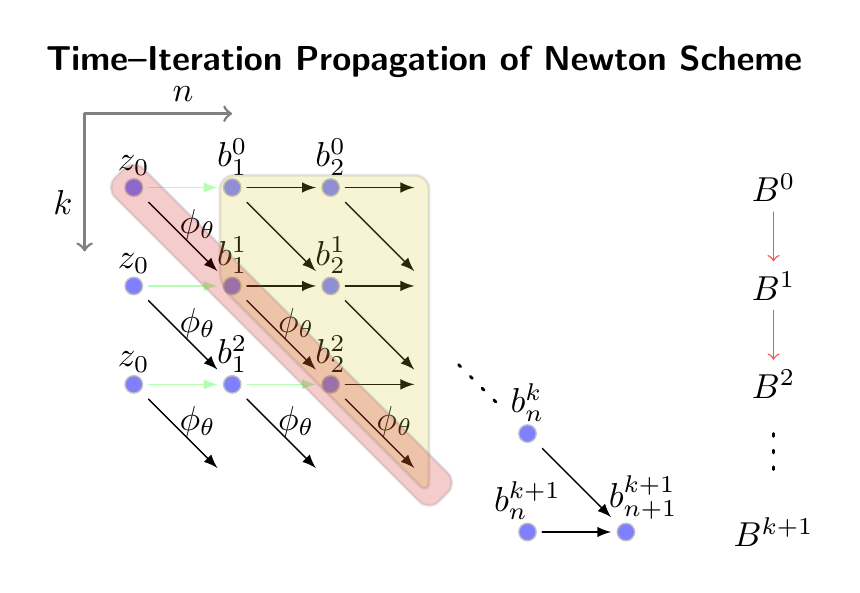}
    \end{center}
    \vspace{-9mm}
    \caption{\footnotesize Propagation in $k$ and $n$ of the Newton iteration \eqref{eq:7}. The intertwining between the updates in $n$ and $k$ leads to the finite step convergence result. In fact, by setting $b_0^0=z_0$, we see how the correcting term $b_n^{k+1}-b_n^k$ multiplying the flow sentitivity $\sD\phi_{\theta,n}$ progressively nullifies at the same rate in $n$ and $k$. As a result, the exact sequential solution of the IVP \eqref{eq:1} unfolds on the diagonal $k=n$ and the only \textit{active} part of the algorithm is the one above the diagonal (highlighted in {\color{yellow!80!black} yellow}).}
    \label{fig:newton}
\end{wrapfigure}

\vspace*{-4.5mm}
\paragraph{Finite-step convergence} Iteration \eqref{eq:7} exhibits convergence to the exact solution of the IVP \eqref{eq:1} in $N-1$ steps \cite[Theorem 2.3]{gander2018time}. In particular, given perfect integration of the sub--IVPs, $b_n^k$ coincides with the exact solution $\phi_\theta (z_0, t_0, t_n)$ from iteration index $k=n$ onward, i.e. at iteration $k$ only the last $N-k$ shooting parameters are actually updated. Thus, the computational and memory footprint of the method diminishes with the number of iterations. This result can be  visualized in the graphical representation of iteration \eqref{eq:7} in Figure \ref{fig:newt_common} while further details are discussed in Appendix \ref{asec:finite_step}.

\paragraph{Numerical implementation}
Practical implementation of the Newton iteration \eqref{eq:7} requires an ODE solver to approximate the flows $\phi_{\theta, n}(b_n^k)$ and an algorithm to compute their sensitivities w.r.t. $b_n^k$. Besides direct application of AD, we show an efficient alternative to obtain all $\sD\phi_{\theta,n}$ in parallel alongside the flows, with a single call of the ODE solver.

\paragraph{Efficient exact sensitivities} %
Differentiating through the steps of the forward numerical ODE solver using reverse--mode AD is straightforward, but incurs in high memory cost, additional computation to \textit{unroll} the solver steps and introduces further numerical error on $\sD\phi_{\theta,n}$. Even though the memory footprint might be mitigated by applying the \textit{adjoint} method \citep{pontryagin1962mathematical}, this still requires to solve backward the $N-k$ adjoint ODEs and sub--IVPs \eqref{eq:2}, at each iteration $k$. We leverage \textit{forward} sensitivity analysis to compute $\sD\phi_{\theta,n}$ alongside $\phi_{\theta,n}$ in a single call of the ODE solver. This approach, which might be considered as the continuous variant of forward--mode AD, scales quadratically with $n_z$, has low memory cost, and explicitly controls numerical error.
\begin{restatable}[Forward Sensitivity \citep{khalil2002nonlinear}]{proposition}{PROPfsens}\label{prop:fsens}
    Let $\phi_\theta(z, s, t)$ be the solution of \eqref{eq:1}. Then, $v(t) = \sD\phi_{\theta}(z,s,t)$ satisfies the linear matrix--valued differential equations
    \[
        \begin{aligned}
            \dot v(t) = \sD f_\theta (t, z(t))v(t),~~~v(s) = \Id_{n_z}~~~\text{where}~\sD f_\theta~\text{denotes}~\partial f_\theta/\partial z.
        \end{aligned}
    \]
\end{restatable}
Therefore, at iteration $k$ all $\sD\phi_{\theta,n}(b_n^k)$ can be computed in parallel while performing the forward integration of the $N-k$ IVPs \eqref{eq:2} and their forward sensitivities, i.e.
\[
    {\tt ForwardSensitivity:~} \left\{b_n^k\mapsto \left(\phi_{\theta, n},\sD\phi_{\theta, n}\right)\right\}_{k<n\leq N}
\]
which enables full \textit{vectorization} of Jacobian--matrix products between $\partial f_\theta / \partial z$ and $v$ as well as maximizing re--utilization of vector field evaluations. {Detailed derivations are provided in Appendix \ref{asec:fw_sens}. Appendix \ref{asec:fsens_soft} analyzes practical considerations and software implementation of the algorithm}.
\paragraph{Zero--order approximate iteration}
In high state dimension regimes, the quadratic memory scaling of the forward sensitivity method might be infeasible. If this is the case, a zero--order approximation of the Newton iteration preserving the finite--step converge property can be employed: the \textit{parareal} method \cite{lions2001resolution}.
From the Taylor expansion of $\phi_{\theta, n}(b_n^{k+1})$ around $b_n^k$
\[
    \begin{aligned}
        \phi_{\theta, n}(b_n^{k+1}) &= \phi_{\theta, n}(b_n^k) + \sD\phi_{\theta,n}(b_n^k)\left(b_n^{k+1}-b_n^k\right) + o\left(\|b_n^{k+1}-b_n^k\|_2^2\right),
    \end{aligned}
\]
we have the following approximant for the correction term of \eqref{eq:7}
\begin{equation}\label{eq:approx_jac}
    \begin{aligned}
    \sD\phi_{\theta,n}(b_n^k)\left(b_n^{k+1}-b_n^k\right) \approx \phi_{\theta, n}(b_n^{k+1})-\phi_{\theta, n}(b_n^k).
    \end{aligned}
\end{equation}
\textit{Parareal} computes the RHS of \eqref{eq:approx_jac} by \textit{coarse}\footnote{e.g. few steps of a low--order ODE solver} numerical solutions $\psi_{\theta, n}(b_n^k),~\psi_{\theta, n}(b_n^{k+1})$ of $\phi_{\theta, n}(b_n^k),~\phi_{\theta, n}(b_n^{k+1})$, leading to the forward iteration,
\[
    b_{n+1}^{b+1} = \phi_{\theta, n}(b_n^k) + \psi_{\theta, n}(b_n^{k+1}) - \psi_{\theta, n}(b_n^k).
\]
\subsection{Properties of MSLs}\label{sec:3.2}
\paragraph{Differentiating through MSL}
Computing loss gradients through MSLs can be performed by directly back--propagating through the steps of the numerical solver via reverse--mode AD.

A memory efficient alternative is to apply the sequential adjoint method to the underlying Neural ODE. 
In particular, consider a loss functions computed independently with the values of different shooting parameters, $L(x, B^*, \theta) = \sum_{n=1}^N c_\theta(x, b^*_n)$. The adjoint gradient for the MSL is then given by
\[
    \nabla_\theta L = \int_0^T\lambda^\top(t) \nabla_\theta f_\theta(t, z(t))\dd t
\]
where the Lagrange multiplier $\lambda(t)$ satisfies a backward piecewise--continuous linear ODE 
\[
    \begin{aligned}
        \dot \lambda(t) &= - \sD f_\theta(t, z(t))\lambda(t) && \text{if }t\in[t_n,t_{n+1})\\
        \lambda^- (t_n) &= \lambda(t_n) + \nabla_{b}^\top c_\theta(x, b_n)~~&&\lambda(T) = \nabla_{b}^\top c_\theta(x, b_N)
    \end{aligned}
\]
The adjoint method typically requires the IVP \eqref{eq:1} to be solved backward alongside $\lambda$ to retrieve the value of $z(t)$ needed to compute the Jacobians $\sD f_\theta$ and $\nabla_\theta f_\theta$. This step introduces additional errors on the final gradients: numerical errors accumulated on $b_N^* \approx z(T)$ during forward pass, propagate to the gradients and sum up with errors on the backward integration of \eqref{eq:1}.

\begin{figure*}[t]
    \centering
    \includegraphics[width=\linewidth]{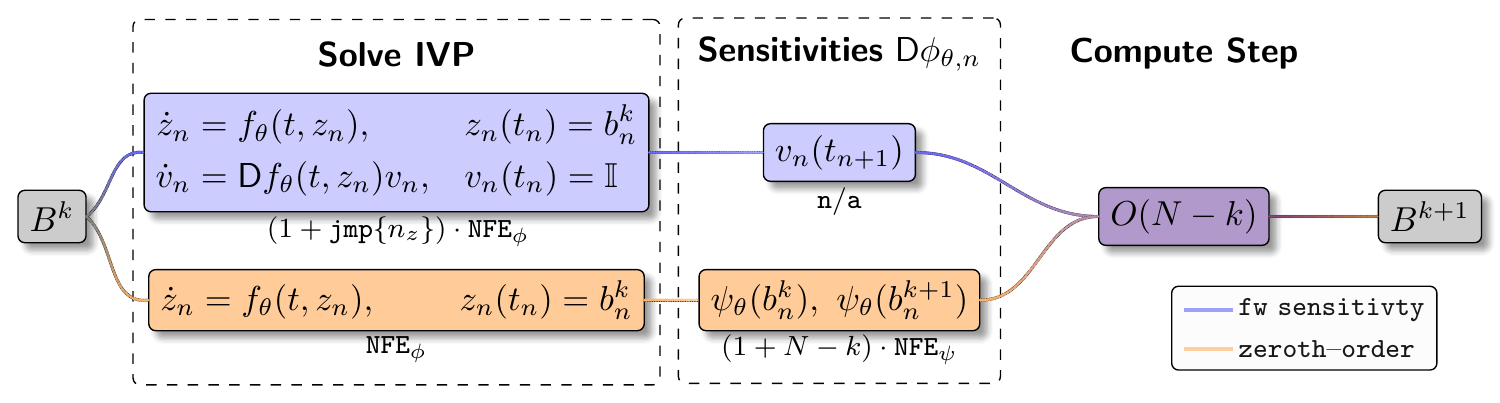}
    \vspace{-7mm}
    \caption{\footnotesize Single iteration computational \textit{span} \citep{mccool2012structured} in MSL. We normalize to $1$ the cost of evaluating $f_\theta$. The $N-k$ sub--IVPs are solved in parallel in their sub--intervals, thus requiring a minimum span ${\tt NFE_\phi}$. \textit{Forward sensitivity} introduces Jacobian--matrix products costs amounting to $\tt jmp$, which can be further parallelized into $\tt jvp$s.}
    \label{fig:forward_cost}
    \vspace{-3mm}
\end{figure*}

Here we take a different, more robust direction by interpolating the shooting parameters and drop the integration of \eqref{eq:1} during the backward pass. The values of the shooting parameters retrieved by the forward pass of MSLs are solution points of the IVP \eqref{eq:1}, i.e. $b_n^* = \phi(z_0, t_0, t_n)$ (up to the forward numerical solver tolerances). On this assumption, we construct a $\tt cubic~spline$ interpolation $\hat z(t)$ of the shooting parameters $b_n^*$ and we query it during the integration of $\lambda$ to compute the Jacobians of $f_\theta$. Further results on back--propagation of MSLs are provided in Appendix \ref{asec:backward}. Appendix \ref{asec:backprop_impl} practical aspects of the backward model alongside software implementation of the \textit{interpolated} adjoint.
\begin{figure*}[t]
    \vspace*{-5mm}
    \centering
    \includegraphics{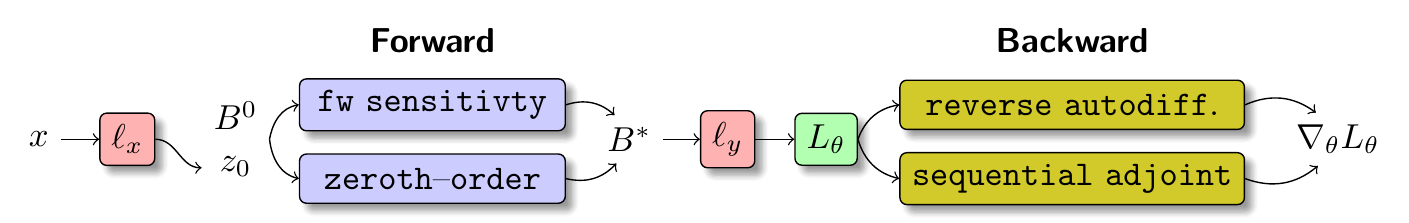}
    \vspace*{-5mm}
    \caption{\footnotesize Scheme of the forward--backward pass of MSL. After applying the input map $\ell_x$ to the input $x$ and choosing initial shooting parameters $B^0$, the forward pass is iteratively computed with one of the numerical schemes described in Sec. \ref{sec:3.1} which, in turn, makes use of some ODE solver to compute $\phi_{\theta,n}$ in parallel, at each step. Once the output $y$ and the loss are computed computed by applying $\ell_y$ and $L_\theta$ to $B^*$, the loss gradients can be computed by standard adjoint methods or reverse--mode automatic differentiation.}
    \label{fig:summary}
\end{figure*}
\paragraph{One-step inference: fixed point tracking}
Consider training a MSL to minimize a twice--differentiable loss function $L(x, B^*, \theta)$ with Lipschitz constant $m_L^\theta$ through the gradient descent iteration
\[
    \theta_{p+1} = \theta_{p} - \eta_p\nabla_\theta L(x, B^*_p, \theta_p)
\]
where $\eta_p$ is a positive learning rate and $B^*_p$ is the exact root of the matching function $g_\theta$ computed with parameters $\theta_p$ (i.e. the exact solution of the IVP \eqref{eq:1} at the boundary points). Due to Lipschitzness of $L$, we have the following uniform bound on the variation of the parameters across training iterations
\[
    \|\theta_{p+1} - \theta_p\|_2\leq\eta_p m_L^\theta.
\]
If we also assume $\gamma_\theta$ to be Lipschitz continuous w.r.t $z$ and $\theta$ with constants $m_\gamma^\theta, ~m_\gamma^z$ and differentiable w.r.t. $\theta$ we can obtain the variation of the fixed point $B^*$ to small changes in the model parameters by linearizing solutions around $\theta_p$
\[
    B^*_{p+1} - B^*_{p} = [\theta_{p+1} - \theta_p]\frac{\partial\gamma_{\theta_{p}}(B^*_p,z_0)}{\partial\theta} + o(\|\theta_{p+1} - \theta_p\|^2_2)
\]
to obtain the uniform bound
\[
    \|B^*_{p+1} - B^*_{p}\|_2 \leq \eta_p m_L^\theta m_\gamma^\theta + o(\|\theta_{p+1} - \theta_p\|^2_2).
\]
Having a bounded variation on the solutions of the MSL for small changes of the model parameters $\theta$, we might think of \textit{recycling} the previous shooting parameters $B^*_p$ as an initial guess for the direct Newton algorithms in the forward pass succeeding the gradient descent update. We show that by choosing a sufficiently small learning rate $\eta_p$ one Newton iteration can be sufficient to track the true value of $B^*$ during training. In particular, the following bound can be obtained.
\begin{restatable}[Quadratic fixed-point tracking]{theorem}{THtrack}\label{th:track}% 
    If $f_\theta$ is twice continuously differentiable in $z$ then 
    \begin{equation}
        \|B^*_{p+1} - \bar B^*_{p}\|_2 \leq M\eta_p^2 
    \end{equation}
    for some $M>0$. $\bar B^{*}_{p}$ is the result of one Newton iteration applied to $B^*_p$.
\end{restatable}
The proof, reported in Appendix \ref{asec:th_track}, relies on the quadratic converge of Newton method. {The quadratic dependence of the tracking error bound on $\eta_k$ allows use of typical learning rates for standard gradient based optimizers} to keep the error under control. In this way, we can turn the \textit{implicit} learning problem into an explicit one where the  implicit inference pass reduces to one Newton iteration. This approach leads to the following training dynamics:
\[
    \begin{aligned}
        \theta &\leftarrow \theta - \eta \nabla_\theta L(x, B^*, \theta)\\
        B^* &\leftarrow {\tt apply}\{\eqref{eq:7},B^*\} 
    \end{aligned}
\]
We note that the main limitation of this method is the assumption on input $x$ to be constant across training iterations (i.e. the initial condition $z_0$ is constant as well). If the input changes during the training (e.g. under mini-batch SGD training regime), the solutions of the IVP \eqref{eq:1} and thus its corresponding shooting parameters may drastically change with $x$ even for small learning rates.
%\clearpage
%
\paragraph{Numerical scaling}
Each class of MSL outlined in Section \ref{sec:3.1} is equipped with unique computational scaling properties. In the following, we denote with ${\tt NFE_\phi}$ the total number of vector field $f_\theta$ evaluations done, in parallel across shooting parameters, in a single sub--interval $[t_n,t_{n+1}]$. Similarly, ${\tt NFE_\psi}$ indicates the function evaluations required by the \textit{coarse} solver used for \textit{parareal} approximations. Here, we set out to investigate the computational signature of MSLs as parallel algorithms. To this end, we decompose a single MSL iteration into two core steps: solving the IVPs across sub--intervals and computing sensitivies $\sD \phi$ (or their approximation). Figure \ref{fig:forward_cost} provides a summary of the \textit{algorithmic span}\footnote{Longest sequential cost, in terms of computational primitives, that is not parallelizable due to problem--specific dependencies. A specific example for MSLs are the sequential ${\tt NFE_\phi}$ calls required by the sequential ODE solver for each sub--interval.} of a single MSL iteration as a function of number of Jacobian vector products ({\tt jvp}), Jacobian matrix products ({\tt jmp}), and vector field evaluations ({\tt NFE}).

{\color{blue!70}{\tt Fw sensitivity} MSL} frontloads the cost of computing $\sD\phi_{\theta,n}$ by solving the forward sensitivity ODEs of Proposition \ref{prop:fsens} alongside the evaluation of $\gamma_{\theta}$. Forward sensitivity equations involve a {\tt jmp}, which can be optionally further parallelized as $n_z$ {\tt jvp}s by paying a memory overhead. Once sensitivies have been obtained along with $\gamma_\theta$, no additional computation needs to take place until application of the shooting parameter update formula. The forward sensitivity approach thus enjoys the highest degree of time--parallelizability at a larger memory cost. 

{\color{orange!70}Zeroth--order MSL} computes $\gamma_\theta$ via a total of ${\tt NFE}_\phi$ evaluations $f_\theta$, parallelized across sub--intervals. The cheaper IVP solution in both memory and compute is however counterbalanced during calculation of the sensitivies, as this MSL approach approximates the sensitivities $\sD\phi_{\theta, n}$ by a zeroth--order update requiring $N-k$ sequential calls to a coarse solver. 

The analysis of MSL backpropagation scaling is straightforward, as sequential adjoints for MSLs mirror standard sensitivity techniques for Neural ODEs in both compute and memory footprints. Alternatively, {\tt AD} can be utilized to backpropagate through the operations of the forward pass methods in use. This approach introduces a non--costant memory footprint which scales in the number of forward iterations and thus depth of computational graph.

%% file: Chapters/4_Applications.tex
\section{Applications}
\subsection{Variational MSL}

Let $x:\R\rightarrow\R^{n_x}$, be an observable of some continuous--time process and let $X=\{x_{-M},\dots, x_0,\dots, x_N\}\in\R^{(M+N+1)\times n_x}$ be a sequence of observations of $x(t)$ at time instants $t_{-M}<\cdots<t_0<\cdots<t_{N}$. We seek a model able to predict $x_1,\dots,x_N$ given past observations $x_{-M},\dots,x_0$, equivalent to approximating the conditional distribution 
$p(x_{1:N}|x_{-M:0})$. To this end we introduce \textit{variational} MSLs ($v$MSLs) as the following latent variable model:
\begin{equation*}
    \begin{aligned}
        \left(\mu, \Sigma\right) &= \cE_\omega(x_{-M:0})~~~~~~~~{\color{gray!70!blue}\text{Encoder}~~\cE_\omega}\\
q_\omega(z_0|x_{-M:0})&=\cN(\mu, \Sigma)~~~~~~~~~~~~{\color{gray!70!blue}\text{Approx. Posterior}}\\
        z_0 &\sim q_\omega(z_0|x_{-M:0})~~~~{\color{gray!70!blue}\text{Reparametrization}}%
    \end{aligned}
    ~~~\vline~~~
    \begin{aligned}
        B^* : g_\theta(B^*, z_0) &= \0 ~~~~~~~~~~~{\color{gray!70!red} \text{Decoder}~~\cD_\theta}\\
        \hat x_1,\dots,\hat{x}_N& =\ell(B^*) ~~~~{\color{gray!70!red}\text{Readout}~~\ell}\\
    \end{aligned}
\end{equation*}
Once trained, such model can be also used to generate new realistic sequences of the observable $x(t)$ by querying the decoder network at a desired $z_0$.
$v$MSLs are designed to scale data generation to longer sequences, exploiting wherever possible parallel computation in time in both {\color{gray!70!blue} encoder} as well as {\color{gray!70!red}decoder} modules.
The structure of $\cE_{\omega}$ is designed to leverage modern advances in representation learning for time--series via temporal convolutions (TCNs) or attention operators \citep{vaswani2017attention} to offer a higher degree of parallelizability compared to sequential encoders e.g RNNs, ODE-RNNs \citep{NEURIPS2019_42a6845a} or Neural CDEs \citep{kidger2020neural}. This, in turn, allows the encoder to match the decoder in efficiency, avoiding unnecessary bottlenecks. 
The decoder $\cD_{\theta}$ is composed of a MSL which is tasked to unroll the generated trajectory in latent space. %
$v$MSLs are trained via traditional likelihood methods. The iterative optimization problem can be cast as the maximization of an evidence lower bound (${\tt ELBO}$): 
\begin{equation*}
        \min_{(\theta, \omega)}\bE_{z_0\sim q_{\omega}(z_0|x_{-M:0})}\sum_{n=1}^N\log p_n(\hat{x}_n) - {\tt KL}(q_{\omega}||\mathcal{N}(\0,\Id))
\end{equation*}
\vspace{-3mm}
with $p_n(\hat{x}_n) = \mathcal{N}(x_n, \sigma_n)$ and the standard deviations $\sigma_n$ are left as a hyperparameters.

\begin{wrapfigure}[14]{l}{0.42\textwidth}
    \vspace*{-3mm}
    \centering
    \includegraphics[width=0.42\textwidth]{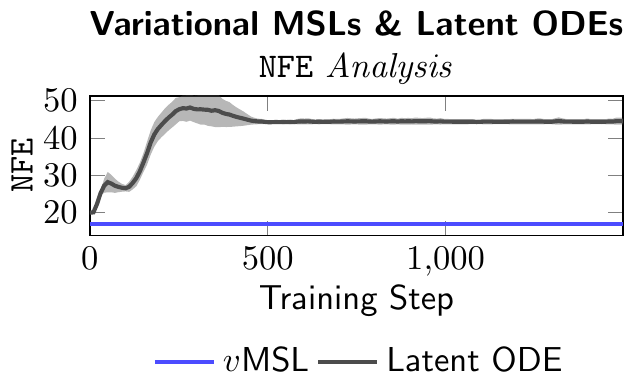}
    \vspace*{-7.5mm}
    \caption{\footnotesize Mean and standard deviation of NFEs during vMSL and Latent Neural ODE across training trials. vMSLs require $60\%$ less NFEs during both training and inference.}
    \label{fig:vmsl_nfe}
    \vspace*{-5mm}
\end{wrapfigure}
\paragraph{Sequence generation under noisy observations}
We apply $v$MSLs on a sequence generation task given trajectories corrupted by state--correlated noise. We consider a baseline latent model sharing the same overall architecture as $v$MSLs, with a Neural ODE as decoder. In particular, the Neural ODE is solved via the {$\tt dopri5$} solver with absolute and relative tolerances set to $10^{-4}$, whereas the $v$MSL decoder is an instance of {\color{blue!70}{\tt fw sensitivity} MSL}. The encoder for both models is comprised of two layers of \textit{temporal convolutions} (TCNs). All decoders unroll the trajectories directly in output space without additional readout networks. The validation on sample quality is then performed by inspection of the learned vector fields and the error against the nominal across the entire state--space. The proposed model obtains equivalent results as the baseline at a significantly cheaper computational cost. As shown in Figure \ref{fig:vmsl_nfe}, $v$MSLs require $60\%$ less NFEs for a single training iteration as well as for sample generation, achieving results comparable to standard Latent Neural ODEs \citep{NEURIPS2019_42a6845a}. %
We report further details and results in Appendix \ref{asec:exp_vml}.
\subsection{Neural Optimal Control}

\begin{wrapfigure}[21]{r}{0.51\textwidth}
    \vspace*{-15mm}
    \centering
    \includegraphics{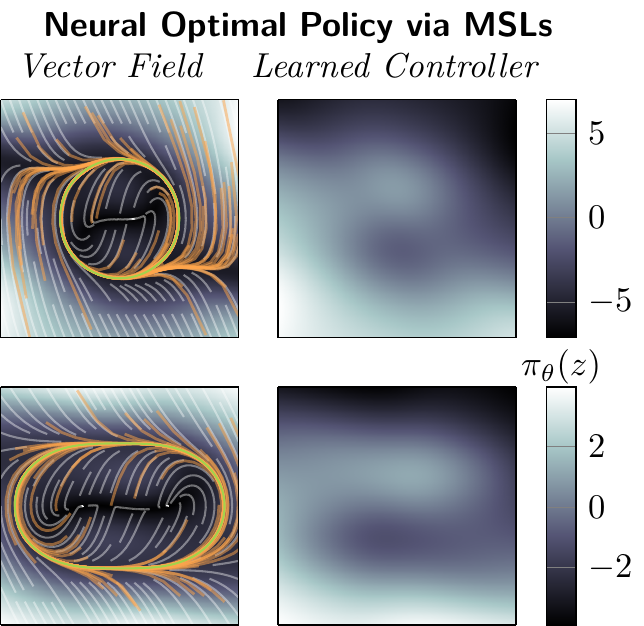}
    \vspace*{-1mm}
    \caption{\footnotesize [Left] Closed--loop vector fields and trajectories corresponding to the $u_\theta$-controlled MSL. [Right] Learned controller $u_\theta(z)$ ($z\in\R^2$) for the two different desired limit cycles. Although, the inference of the all trajectory is performed with just two steps of {\tt RK4} (8 {\tt NFE}), the initial accuracy of {\tt dopri5} ($>3000$ {\tt NFE}) is preserved throughout training.}
    \label{fig:cmsl_stream}
\end{wrapfigure}
Beyond sequence generation, the proposed framework can be applied to \textit{optimal control}. Here we can fully exploit the drastic computational advantages of MSLs. In fact, we leverage on the natural assumption of finiteness of initial conditions ${z_0}$ where the controlled system (or \textit{plant}) is initialized to verify the result of Th. \ref{th:track}.
Let us consider a controlled dynamical system
\begin{equation}\label{eq:8}
        \dot z(t) = f(t, z(t), \pi_\theta(t, z)),~~~
        z(0) =z_0%
\end{equation}
with a parametrized policy $\pi_\theta : t,z\mapsto \pi_\theta(t, z)$ and initial conditions $z_0$ ranging in a finite set $Z_0=\{z_0^j\}_j$. We consider the problems of stabilizing a low--dimensional \textit{limit cycle} and deriving an optimal boundary policy for a linear PDE discretized as a $200$--dimensional ODE.

\begin{wrapfigure}[19]{r}{0.51\textwidth}
    \centering
    \vspace*{-7mm}
    \includegraphics[width=0.51\textwidth]{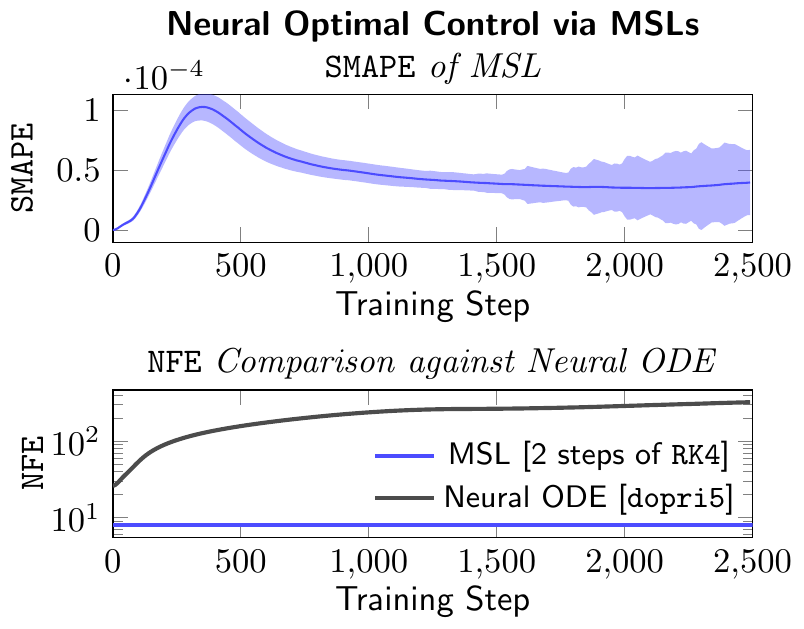}
    \vspace*{-7.5mm}
    \caption{\footnotesize \textit{Symmetric Mean Average Percentage Error} (SMAPE) between solutions of the controlled systems obtained by MSLs and nominal. Compared to Neural ODEs, MSLs solve the optimal control problem with NFE savings of several orders of magnitude by carrying forward their solution across training iterations.}
    \label{fig:cmsl_nfe}
\end{wrapfigure}

\paragraph{Limit cycle stabilization}
We consider a \textit{stabilization} task where, given a desired closed curve $S_d=\{z\in\cZ:s_d(z)=0\}$, we minimize the 1-norm between the given curve and the MSL solution of \eqref{eq:8} across the timestamps%
as well as the \textit{control effort} $|\pi_\theta|$. We verify the approach on a one degree--of--freedom mechanical system, aiming with closed curves $S_d$ of various shapes. Following the assumptions on fixed point tracking and slow--varying--flows of Th. \ref{th:track}, we initialize $B_j^0$ by {$\tt dopri5$} adaptive--step solver set with tolerances $10^{-8}$. Then, at each training iteration, we perform inference with a single parallel {$\tt rk4$} step for each sub--interval $[t_n, t_{n+1}]$ followed by a single Newton update. Figure \ref{fig:cmsl_stream} shows the learned vector fields and controller, confirming a successful system stabilization of the system to different types of closed curves. 
We compare with a range of baseline Neural ODEs, solved via {$\tt rk4$} and {$\tt dopri5$}.
Training of the controller via MSLs is achieved with orders of magnitude \textit{less} wall--clock time and NFEs. Figure \ref{fig:cmsl_nfe} shows the difference in NFEs w.r.t. {$\tt dopri5$}. 

We further provide \textit{Symmetric Mean Average Percentage Error} (SMAPE) measurements between trajectories obtained via MSLs and an adaptive--step solver. MSLs initialized with recycled solutions are able to track the nominal trajectories across the entire training process. 
Additional details on the experimental setup, including wall--clock time comparisons with {$\tt rk4$} and ${\tt dopri5}$ baseline Neural ODEs  is are provided in Appendix \ref{asec:exp_ctrl}.
\paragraph{Neural Boundary Control of the Timoshenko Beam}
We further show how MSLs can be scaled to high--dimensional regimes by tackling the optimal boundary control problem for a linear partial differential equation. In particular, we consider the \textit{Timoshenko beam} \citep{macchelli2004modeling} model in Hamiltonian form. We derive formalize the boundary control problem and obtain a structure--preserving spectral discretization yielding a 200--dimensional Hamiltonian ODE. 

\begin{wrapfigure}[14]{l}{0.45\textwidth}
    \vspace*{-1.5mm}
    \centering
    \includegraphics[scale=.95]{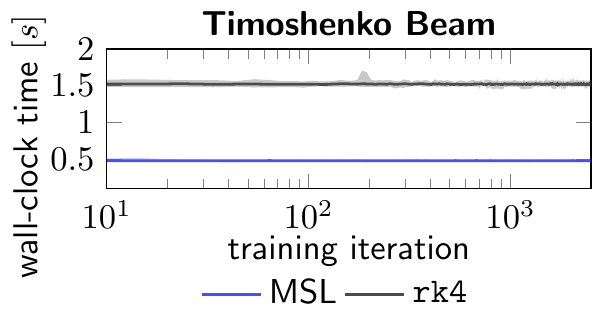}
    \vspace*{-2mm}
\caption{\footnotesize Mean and standard deviation of wall--clock time per training iteration for MSL and Neural ODE across training trials. MSLs require are three times faster than the sequential ${\tt rk4}$ with same accuracy (step size).}
    \label{fig:wall_clock_pde}
\end{wrapfigure}

We parameterize the boundary control policy with a multi--layer perceptron taking as input (control feedback) the 200--dimensional discrete state. We train the model in similar setting to the previous example having the MSL equipped with {\color{blue!70} {\tt fw sensitivity}} and one step of {\tt rk4} for the parallel integration. We compare the training wall--clock time with a Neural ODE solved by sequential ${\tt rk4}$. The resulting speed up of MSL with forward sensitivity is three time faster than the baseline Neural ODE proving that the proposed method is able to scale to high--dimensional regimes. We include a formal treatment of the boundary control problem in Appendix \ref{asec:pde} while further experimental details are provided in Appendix \ref{asec:exp_pde}.
\subsection{Fast Neural CDEs for Time Series Classification}

\begin{wrapfigure}[10]{r}{0.45\textwidth}
    \centering
    \vspace{-8mm}
    \includegraphics[scale=0.95]{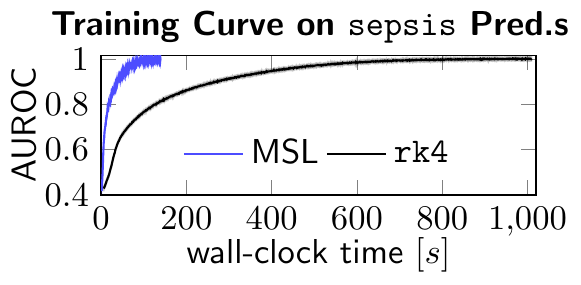}
    \vspace{-2mm}
    \caption{\footnotesize Mean and standard deviation of {\tt AUROC} during training of MSLs and baseline Neural CDEs on sepsis prediction.}
    \label{fig:cde}
\end{wrapfigure}

To further verify the broad range of applicability of MSLs, we apply them to time series classification as faster alternatives to \textit{neural controlled differential equations} (Neural CDEs) \citep{kidger2020neural}. 
Here, MSLs remain applicable since Neural CDEs are practically solved as ODEs with a special structure, as described in Appendix \ref{asec:exp_cde}. We tackle the {$\tt PhysioNet$} 2019 challenge  \citep{goldberger2000physiobank} on {$\tt sepsis$} prediction, following the exact experimental procedure described by \cite{kidger2020neural}, including hyperparameters and Neural CDE architectures. However, we train all models on the full dataset to enable application of the fixed point tracking technique for MSLs\footnote{We note that the training for all models has been performed on a single \textsc{NVIDIA RTX A6000} with $48\tt Gb$ of graphic memory.}. Figure \ref{fig:cde} visualized training convergence of {\color{orange!70} zeroth--order MSL} Neural CDEs and the baseline Neural CDE solved with rk4 as in the original paper. Everything else being equal, including architecture and backpropagation via sequential adjoints, MSL Neural CDEs converge with total wall--clock time one order of magnitude smaller than the baseline.

%% file: Chapters/5_Discussion.tex
\section{Related Work}
\paragraph{Parallel--in--time integration \& multiple shooting}
MSLs belong to the framework of time--parallel integration algorithms. The study of these methods is relatively recent, with seminal work in the 60s \citep{nievergelt1964parallel}.  The \textit{multiple shooting} formulation of time--parallel integration, see e.g. \citep{bellen1989parallel} or \citep{chartier1993parallel}, finally lead to the modern algorithmic form using the Newton iteration reported in \eqref{eq:7}. \textit{Parareal} \citep{lions2001resolution} has been successively introduced as a cheaper approximated solution of multiple shooting problem, rapidly spreading across application domains, e.g. optimal control of partial differential equations \citep{maday2002parareal}. We refer to \citep{gander201550,gander2018time} as excellent introduction to the topic.
We also note recent work \citep{vialard2020shooting} introducing \textit{single shooting} terminology for Neural ODEs \citep{NEURIPS2018_69386f6b}, albeit in the unrelated context of learning time--varying parameters. 

\paragraph{Time--parallelization in neural models}
In the pursuit for increased efficiency, several works have proposed approaches to parallelization across time or \textit{depth} in neural networks.
\citep{gunther2020layer,kirby2020layer,sun2020penalty} use multigrid and penalty methods to achieve speedups in ResNets. \cite{meng2020ppinn} proposed a parareal variant of Physics--informed neural networks (PINNs) for PDEs. \cite{zhuang2021multiple} uses a penalty--variant of multiple shooting with adjoint sensitivity for parameter estimation in the medical domain. Solving the boundary value problems with a regularization term, however, is not guaranteed to converge to a continuous solution. The method of \cite{zhuang2021multiple} further optimizes its parameters in a full--batch regime, where application of \eqref{th:track} achieves drastic speedups while preserving convergence guarantees. Recent theoretical work \citep{lorin2020derivation} has applied \textit{parareal} methods to Neural ODEs. However, their analysis is limited to the theoretical computational complexity setting and does not involve multiple shooting methods nor derives its implicit differentiation formula. 

In contrast our objective is to introduce a novel class of implicit time--parallel models, and to validate their computational efficiency across settings.% \ref{th:1}
\section{Conclusion}
This work introduces differentiable \textit{Multiple Shooting Layers} (MSLs), a parallel--in--time alternative to neural differential equations. MSLs seek solutions of differential equations via parallel application of root finding methods across solution subintervals. Here, we analyze several model variants, further proving a fixed point tracking property that introduces drastic speedups in full--batch training. The proposed approach is validated on different tasks: as generative models, MSLs are shown to achieve same task performance as Neural ODE baselines with 60\% less NFEs, whereas they are shown to offer several orders of magnitude faster in optimal control tasks.
\vfill
\begin{center}\small
    \textit{Remarkably few methods have been proposed for parallel integration of ODEs. In part this is because the problems do not have much natural parallelism. (\citealp{gear1988parallel})}
\end{center}
\clearpage

%% file: Appendix/Appendix_A.tex
\clearpage
\section{Proofs}
\subsection{Proof of Theorem \ref{th:track}}\label{asec:th_track}
\THtrack*
\proof
    For compactness, we neglect the dependence of $\gamma_\theta, ~g_\theta$ on $z_0$ and we write $\gamma(B,\theta) = \gamma_\theta(B, z_0), g(B,\theta) = g_\theta(B, z_0)$. Recalling that by definition of $B^*_{p+1}$ it holds
    \begin{equation}
        g(B^*_{p+1}, \theta_{p+1}) = B^*_{p+1} - \gamma(B^*_{p+1}, \theta_{p+1}) = \0,
    \end{equation}
    we write the 2-jet of the matching equation at $B^*_p$,
    \begin{equation}\label{eq:proof_1}
        \begin{aligned}
            g(B^*_{p+1}, \theta_{p+1}) &=  g(B^*_{p}, \theta_{p+1}) + \sD g(B^*_{p}, \theta_{p+1})\left[B^*_{p+1} - B^*_{p}\right]\\
            &+ \frac{1}{2}\sD^2 g(B^*_{p}, \theta_{p+1})\left[B^*_{p+1} - B^*_{p} \right]^{\otimes 2} + o(\|B^*_{p+1} - B^*_{p}\|^3_2)
        \end{aligned}
    \end{equation}
    where $\sD g,~\sD^2 g$ can be computed thanks to the assumptions on differentiability of $f_\theta$. From the Newton iteration we have that  
    \begin{equation}\label{eq:proof_2}
        \begin{aligned}
            &\bar B^*_{p} - B^*_{p} = -[\sD g(B^*_{p},\theta_{p+1})]^{-1}g(B^*_{p},\theta_{p+1})\\
            \Leftrightarrow~~~&\0 = g(B^*_{p},\theta_{p+1}) + \sD g(B^*_{p},\theta_{p+1})\left[\bar B^*_{p} - B^*_{p}\right]
        \end{aligned}
    \end{equation}
    Using $\0 = g(B^*_{p+1}, \theta_{p+1})$, we subtract \eqref{eq:proof_2} from \eqref{eq:proof_1} yielding
    \begin{equation}\label{eq:proof_3}
        \begin{aligned}
            \0 &=  \cancel{g(B^*_{p}, \theta_{p+1})} + \sD g(B^*_{p}, \theta_{p+1})\left[B^*_{p+1} - B^*_{p}\right] - \cancel{g(B^*_{p},\theta_{p+1})} - \sD g(B^*_{p},\theta_{p+1})\left[\bar B^*_{p} - B^*_{p}\right]\\
            &+ \frac{1}{2}\sD^2 g(B^*_{p}, \theta_{p+1})\left[B^*_{p+1} - B^*_{p} \right]^{\otimes 2} + o(\|B^*_{p+1} - B^*_{p}\|^3_2)\\
            & = \sD g(B^*_{p}, \theta_{p+1})\left[B^*_{p+1} - B^*_{p}\right] - \sD g(B^*_{p},\theta_{p+1})\left[\bar B^*_{p} - B^*_{p}\right]\\
            &+ \frac{1}{2}\sD^2 g(B^*_{p}, \theta_{p+1})\left[B^*_{p+1} - B^*_{p} \right]^{\otimes 2} + o(\|B^*_{p+1} - B^*_{p}\|^3_2)\\
            & = \sD g(B^*_{p}, \theta_{p+1})\left[B^*_{p+1} - \bar B^*_{p}\right] + \frac{1}{2}\sD^2 g(B^*_{p}, \theta_{p+1})\left[B^*_{p+1} - B^*_{p} \right]^{\otimes 2} + o(\|B^*_{p+1} - B^*_{p}\|^3_2).
        \end{aligned}
    \end{equation}
    Being the Jacobian of $g_\theta$
    \[
        \sD g(B^*_{p}, \theta_{p+1}) = \Id_N\otimes \Id_{n_z} - \sD \gamma(B^*_{p}, \theta_{p+1}) = 
        \begin{bmatrix}
            \Id_{n_z} &\times&\times&\times\\
            -\sD\phi_{\theta_{p+1},0}(b_{0,p}^*) & \Id_{n_z}&\times&\times\\
            \times&\ddots & \ddots&\times\\
            \times&\times&-\sD\phi_{\theta_{p+1},N-1}(b_{N-1,p}^*) & \Id_{n_z}\\
        \end{bmatrix}
    \]
    always invertible due to the nilpotency of $\sD\gamma$, we can solve \eqref{eq:proof_3} in terms for $B^*_{p+1} - \bar B^*_{p}$, leading to
    \begin{equation}
        B^*_{p+1} - \bar B^*_{p} = \frac{1}{2}\left[\sD g(B^*_{p}, \theta_{p+1})\right]^{-1}\sD^2 g(B^*_{p}, \theta_{p+1})\left[B^*_{p+1} - B^*_{p} \right]^{\otimes 2} + o(\|B^*_{p+1} - B^*_{p}\|^3_2).
    \end{equation}
    Taking the norm we have
    \begin{equation}
        \begin{aligned}
            \|B^*_{p+1} - \bar B^*_{p}\|_2 &\leq \frac{1}{2}\|[\sD g(B^*_{p}, \theta_{p+1})]^{-1}\|_2\|\sD^2 g(B^*_{p}, \theta_{p+1})\|_2\|B^*_{p+1} - B^*_{p}\|_2^2.
        \end{aligned}
    \end{equation}
    Using 
    \[
         \|B^*_{p+1} - B^*_{p}\|_2 \leq \eta_p m_L^\theta m_\gamma^\theta
    \]
    and $\|\sD^2 g(B^*_{p}, \theta_{p+1})\|_2 = \|\sD^2 \gamma(B^*_{p}, \theta_{p+1})\|_2\leq m^z_{\partial\gamma}$
    we obtain
    \begin{equation}
        \begin{aligned}
            \|B^*_{p+1} - \bar B^*_{p}\|_2 &\leq \frac{1}{2}\eta_p^2 (m_L^\theta m_\gamma^\theta)^2m^z_{\partial\gamma}\|[\sD g(B^*_{p}, \theta_{p+1})]^{-1}\|_2.
        \end{aligned}
    \end{equation}
    Since $R = \sD \gamma(B^*_{p}, \theta_{p+1})$ is a nilpotent matrix then 
    \[
        [\sD g(B^*_{p}, \theta_{p+1})]^{-1} = [\Id_N\otimes\Id_{n_z} - R]^{-1} = \Id_N\otimes\Id_{n_z} + \sum_{n=1}^N R^n
    \]
    and
    \[
        \|[\sD g(B^*_{p}, \theta_{p+1})]^{-1}\|_2 \leq 1 + \sum_{n=1}^N \|R\|_2^n.
    \]
    By Lipsichitz continuity of $\gamma(B^*_{p}, \theta_{p+1})$ we have that 
    \[
        \|R\|_2\leq m_\gamma^z
    \]
    and 
    \[
        \|[\sD g(B^*_{p}, \theta_{p+1})]^{-1}\|_2 \leq 1 + \sum_{n=1}^N (m_\gamma^z)^n
    \]
    By convergence of finite geometric series we obtain 
    \[
        \|[\sD g(B^*_{p}, \theta_{p+1})]^{-1}\|_2 \leq \frac{1 - (m_\gamma^z)^{N+1}}{1 - m_\gamma^z}
    \]
    The final bound on the tracking error norm thus becomes
    \begin{equation}
        \begin{aligned}
            \|B^*_{p+1} - \bar B^*_{p}\|_2 &\leq \frac{1}{2}\eta_p^2 (m_L^\theta m_\gamma^\theta)^2m^z_{\partial\gamma}\frac{1 - (m_\gamma^z)^{N+1}}{1 - m_\gamma^z}.
        \end{aligned}
    \end{equation}
    The proof of the theorem is completed by setting 
    \[
      M >\frac{1}{2} (m_L^\theta m_\gamma^\theta)^2m^z_{\partial\gamma}\frac{1 - (m_\gamma^z)^{N+1}}{1 - m_\gamma^z}
    \]
\endproof
%%%%%%%%%%%%%%%%%%%%%%%%%%%%%%%%%%
\subsection{Proof of Proposition \ref{prop:fsens}}
\PROPfsens*
\proof
The following proof is adapted from \cite[Section 3.3]{khalil2002nonlinear}. If $\phi_\theta(z_0, s, t)$ is a solution of \eqref{eq:1} at time $t$ starting from $z_0$ at time $s$, $s<t;~s, t\in[t_0, t_N]$ then
\begin{equation}
    \phi_\theta(z_0, s, t)  = z(t) = z_0 + \int_s^t f_\theta(\tau, z(\tau))\dd\tau 
\end{equation}
Differentiating under the integral sign w.r.t. $z$ yields
\begin{equation}
    \begin{aligned}
        \sD \phi_\theta(z_0, s, t) = \frac{\dd z(t)}{\dd z_0} &= \frac{\dd z_0}{\dd z_0} + \int_s^t\frac{\partial f_\theta(\tau, z(\tau))}{\partial z(\tau)}\frac{\dd z(\tau)}{\dd z_0}\dd\tau\\
            &= \Id_{n_z} + \int_s^t\frac{\partial f_\theta(\tau, z(\tau))}{\partial z(\tau)}\frac{\dd z(\tau)}{\dd z_0}\dd\tau
    \end{aligned}
\end{equation}
We denote $\sD f_\theta(t, z(t)) = {\partial f_\theta(\tau, z(\tau))}/{\partial z(\tau)}$ and we notice that ${\dd z(\tau)}/{\dd z_0}$ is the flow Jacobian $\sD \phi_\theta(z_0, s, \tau)$ at time $\tau\in[s, t]$. Then, the function $v:[s, t]\rightarrow\R^{n_z \x n_z};~\tau\mapsto\sD \phi_\theta(z_0, s, \tau)$ satisfies
\[
    v(t) = \Id_{n_z} + \int_s^t\sD f(\tau, z(\tau)) v(\tau)\dd \tau
\]
or, in differential form, $v$ satisfies the IVP
\[
    \dot v (\tau) =  \sD f(\tau, z(\tau)) v(\tau),\quad v(s) = \Id_{n_z}. 
\]
\endproof

%% file: Appendix/Appendix_B.tex
\clearpage
\section{Additional Theoretical Results}
\subsection{Finite--Step Convergence}\label{asec:finite_step}
We discuss more rigorously the intuitions on the finite--step convergence of direct Newton methods introduced in the main text. The following results are thoroughly detailed in \citep{gander2018time}. We recall that, by assuming that the first shooting parameter is correctly initialized to $z_0$ and the numerical integration is \textit{exact} (we can perfectly retrieve the sub-flows $\phi_{\theta, n}$), the shooting parameters $b_n^k$ coincides with the exact solution of \eqref{eq:1} from Newton iteraion $k=n$ onward. Formally,
\begin{proposition}[Finite--step convergence]\label{prop:finite_step}
    If $b_0^0 = z_0$, then solution of the Newton iteration \eqref{eq:7} are such that
    \begin{equation}
        k\geq n ~\Rightarrow~b_n^k = \phi_\theta(z_0, t_0, t_n).
    \end{equation}
\end{proposition}
\proof 
The proof is obtained by induction on the shooting parameter index $n$ (\textit{time} direction) and follows from \citep[Theorem 2.3]{gander2018time}. 
\begin{itemize}
    \item[$i.$] ({\color{blue!70!white} {$\tt base~case$}: $n=0$}) For $n=0$,  $b_0^0 = z_0$ by assumption. Moreover, the iteration $\eqref{eq:7}$ yields $b_0^k = b_0^{k+1} = z_0$ for all naturals $k$. 
    \item[$ii.$] ({\color{blue!70!white} {$\tt induction~step$}: $n \rightarrow n+1$}) Suppose that 
    \[
        k\geq n ~\Rightarrow~b_n^k = \phi_\theta(z_0, t_0, t_n).
    \]
    We need to show that 
    \[
        k+1\geq n+1 ~\Rightarrow~b_{n+1}^{k+1} = \phi_\theta(z_0, t_0, t_{n+1}),
    \]
    to conclude the proof by induction. We notice that if we increase $k$ to $k+1$, then $k+1$ is still greater than $n$ yielding $b_n^{k+1} = b_n^k = \phi_\theta(z_0, t_0, t_n)$. Using \eqref{eq:7}, we have
    \[
        \begin{aligned}
            b_{n+1}^{k+1} &= \phi_{\theta,n}(b_n^k) + \sD\phi_{\theta, n}(b_n^k)\left(b_n^{k+1} - b_n^k\right)\\
                          &=  \phi_{\theta,n}(\phi_\theta(z_0, t_0, t_n)) + 0 &&\text{\small by induction hypothesis $b_n^k=b_n^{k+1}=\phi_\theta(z_0, t_0, t_n)$;}\\
                          &= \phi_\theta(z_0, t_0, t_{n+1}) && \text{\small by the flow property of ODE solutions;}
        \end{aligned}       
    \]
    where the induction hypothesis has been used thanks to the fact that $k+1\geq n+1\Rightarrow k\geq n$.
\end{itemize}
\endproof
The above result can be also extended to the zeroth--order (\textit{parareal}) method as follows.
\begin{proposition}[Finite--step convergence w/ zeroth--order Jacobian approximation]\label{prop:finite_step_parareal}
    If $b_0^0 = z_0$, then solution of the approximate Newton iteration 
    \begin{equation}\label{aeq:parareal}
        b_{n+1}^{k+1} = \phi_{\theta,n}(b_n^k) + \psi_{\theta,n}(b_n^{k+1}) - \psi_{\theta,n}(b_n^{k})
    \end{equation}
    are such that
    \begin{equation}
        k\geq n ~\Rightarrow~b_n^k = \phi_\theta(z_0, t_0, t_n).
    \end{equation}
\end{proposition}
\proof 
The proof is identical to the one Proposition \ref{prop:finite_step} where \eqref{aeq:parareal} is used in the induction step and noticing that the correction term $\psi_{\theta,n}(b_n^{k+1}) - \psi_{\theta,n}(b_n^{k})$ nullifies for $k>n$ by induction hypothesis $b_n^k=b_n^{k+1}=\phi_\theta(z_0, t_0, t_n)$.
\endproof
{
Even though Proposition \ref{prop:finite_step} and Proposition \ref{prop:finite_step_parareal} show that the direct Newton method (and its zeroth-order approximation) will always converge to the exact solution of \eqref{eq:1}, full convergence after $N$ iterations is completely useless from a practical perspective. If we suppose to use a \textit{fine solver} $\tilde\phi_{\theta, n}$ to obtain in parallel accurate numerical approximations of the sub--flows $\phi_{\theta,n}$ and we iterate \eqref{eq:7} $N$ times, we will also have executed the parallel integration $N$ times. Thus, one could also just have applied the same fine solver sequentially across the $N$ boundary points $t_n$ with one processing thread and obtain the same result. For this reason, we believe that tracking Theorem \ref{th:track} is a key result to obtain large speedups in the machine learning applications of MSLs.
}%
\subsection{Flows Sensitivities $\sD\phi_{\theta,n}$}\label{asec:fw_sens}
The most computationally demanding stage of the MSL inference is without any doubts the correction term 
\[
    \sD\phi_{\theta,n}(b_n^k)\left(b_n^{k+1} - b_n^k\right)
\]
of the direct Newton iteration $\eqref{eq:7}$. In this paper, we propose to either use the forward sensitivity approach of Proposition \ref{prop:fsens} or to rely on the zeroth--order approximation of parareal. Moreover, we discouraged the use of both reverse--mode AD and backward adjoint sensitivities to compute the full Jacobians $\sD\phi_{\theta,n}$ due to their higher memory or computational cost.

\paragraph{Sensitivities with ${N-k}$ {\tt jvp}s} A common feature among the aforementioned approaches (but the parareal) is that all $\sD\phi_{\theta,n}$ can be computed in parallel at the beginning of each Newton iteration with a single call of the sensitivity routine. An alternative \textit{sequential} approach relies on computing $\sD\phi_{\theta,n}(b_n^k)\left(b_n^{k+1} - b_n^k\right)$ directly as a $\tt jvp$ during each step of \eqref{eq:7}. This method avoids the computation of the full Jacobians at cost of having to call the $\tt jvp$ routine $N-k$ times at each Newton iteration. In such case the only parallel operation performed is the integration of the sub--flows $\phi_{\theta,n}$. Nonetheless, we believe that this direction is worth to be explored in future works.
\subsection{Backward Model of Multiple Shooting Layers}\label{asec:backward}
We show how MSLs can be trained via standard gradient descent techniques where gradients can be either computed by back--propagating through the operations of the forward pass (parallel/memory intensive) or by using the convergence property of direct Newton method and directly apply a interpolated adjoint routine\footnote{Implementation details are provided in Appendix \ref{asec:backprop_impl}} (sequential/memory efficient). Although we believe that these two approaches to backpropagation are sufficient within the scope of this manuscript as they allow for substantial computational speedups and robustness, we hereby report further theoretical considerations on the backward pass of MSLs. A thorough algorithmic and experimental analysis of the following content is a promising research direction for future work.

\paragraph{Implicit differentiation of MSLs} As repeatedly pointed out throughout the paper, MSLs are implicit models and satisfy the implicit relation
\begin{equation}\label{aeq:implicit}
     B^* ~:~ B^* = \gamma_\theta(B^*, z_0).
\end{equation}
It thus make sense to interpret the backward pass of MSLs in an \textit{implicit} sense. In particular, implicit differentiation of the relation \eqref{aeq:implicit} at $B^*$ leads to the following loss gradients.
\vspace*{2mm}
\begin{restatable}[Implicit Gradients]{theorem}{THgrad}\label{th:1}
    Consider a smooth loss function $L_\theta$. It holds
    \begin{equation}\label{eq:4}
        \frac{\dd L_\theta}{\dd\theta} = \frac{\partial L_\theta}{\partial\theta} + \frac{\partial L_\theta}{\partial\ell_y}\frac{\partial \ell_y}{\partial B^*}\left[\Id_{n_z}\otimes\Id_{N} - \sD\gamma_\theta(B^*)\right]^{-1}\frac{\partial \gamma_\theta}{\partial \theta} 
    \end{equation}
    where $\sD\gamma_\theta(B^*, z_0)\in\R^{Nn_z\times Nn_z}$ is the Jacobian of $\gamma_\theta$ computed at $B^*$.
\end{restatable}
\proof
    By application of the chain rule to the MSL forward model \eqref{eq:3} we obtain
    $$    	
    	\frac{\dd L_\theta}{\dd\theta} = \frac{\partial L_\theta}{\partial \theta} + \frac{\partial L_\theta}{\partial \ell_y}\frac{\partial \ell_y}{\partial B^*}\frac{\dd B^*}{\dd \theta}.
    $$
    With
    $$
    	\begin{aligned}
    	g_\theta(B^*) = \0~&\Rightarrow~B^* - \gamma(B^*, z_0) = \0\\
    	&\Leftrightarrow\frac{\partial g_\theta(B^*)}{\partial \theta} +\left[\Id_{n_z}\otimes\Id_{N} - \sD\gamma_\theta(B^*)\right]\frac{\dd B^*}{\dd\theta}=0\\
    	&\Leftrightarrow~\frac{\dd B^*}{\dd\theta} = \left[\Id_{n_z}\otimes\Id_{N} - \sD\gamma_\theta(B^*)\right]^{-1}\frac{\partial \gamma_\theta}{\partial\theta}.
    	\end{aligned}
    $$
    Thus,
    $$
    	\begin{aligned}
    	\frac{\dd L_\theta}{\dd\theta} &= \frac{\partial L_\theta}{\partial\theta}\frac{\partial L_\theta}{\partial \ell_y}\frac{\partial \ell_y}{\partial B^*} \left[\Id_{n_z}\otimes\Id_{N} - \sD\gamma_\theta(B^*)\right]^{-1}\frac{\partial \gamma_\theta}{\partial\theta}
    	\end{aligned}
    $$
    where the Jacobian $\sD\gamma_\theta(B^*)$ is computed as
    \begin{equation*}
        \sD\gamma_\theta(B^*) = 
        \begin{bmatrix}
            \x &\times&\times&\times&\times\\
            \sD\phi_{\theta,0}(b^*_0) & \x&\times&\times&\times\\
            \times&\sD\phi_{\theta,1}(b^*_1) & \x&\times&\times\\\\
            \times&\times&\ddots & \ddots&\times\\\\
            \times&\times&\times&\sD\phi_{\theta,N-1}(b^*_{N-1}) & \x\\
        \end{bmatrix}
    \end{equation*}
    and 
    \begin{equation*}
        \frac{\partial\gamma_\theta}{\partial\theta} =
        \begin{bmatrix}
        	\0_{n_z}\\
        	\dfrac{\partial\phi_{\theta,0}(b_0)}{\partial\theta}\\
        	\vdots\\
        	\dfrac{\partial\phi_{\theta,N-1}(b_{N-1})}{\partial\theta}
        \end{bmatrix}
    \end{equation*}
\endproof
The implicit differentiation routine suggested by Theorem \ref{th:1} presents two terms which appear to be very demanding both memory and computation--wise:
\begin{itemize}
    \item[$(i)$] The inverse Jacobian $\left[\Id_{n_z}\otimes\Id_{N} - \sD\gamma_\theta(B^*)\right]^{-1}$ of the implicit relation;
    \item[$(ii)$] The sub--flows sensitivities to the model parameters $\theta$.
\end{itemize}
In order to retrieve $(i)$ in standard \textit{Deep Equilibrium Models} \cite{bai2019deep}, one should either compute the full--Jacobian at the fixed point via AD and invert it or ``recycle'' its low--rank approximation from the Quasi--Newton method of the forward pass. In the case of MSLs we can take advantage of the special structure of the implicit relation to obtain the exact Jacobian inverse in a computationally efficient manner. In fact, if {\color{blue!70!white}\tt fw--sensitivity} has been used in the forward pass to compute $B^*$, then the sensitivities of the sub--flows computed at the last step $K$ of the Newton iteration $\sD\phi_{\theta, n}(b_n^K)$ can be stored and re--used to construct the Jacobian $\sD\gamma_\theta$.
Further, due to the nilpotency of $\sD\gamma_\theta$ the inverse of the total Jacobian can be retrieved in closed form by the finite matrix power series
\[
    \left[\Id_{n_z}\otimes\Id_{N} - \sD\gamma_\theta(B^*)\right]^{-1} = \Id_{n_z}\otimes\Id_{N} + \sum_{n=1}^N[\sD\gamma_\theta(B^*)]^n.
\]
Finally, $(ii)$ may be indirectly computed with a single {\tt vjp}
\[
    v^\top\frac{\partial\gamma_\theta(B^*, z_0)}{\partial\theta}
\]
with $v^\top$ being a 1 by $Nn_\theta$ row vector defined as
\[
    v^\top = \frac{\partial L_\theta}{\partial\ell_y}\frac{\partial \ell_y}{\partial B^*}\left[\Id_{n_z}\otimes\Id_{N} + \sum_{n=1}^N[\sD\gamma_\theta(B^*)]^n\right]
\]
leading to the implicit cost gradient with a single call of the AD.

%% file: Appendix/Appendix_C.tex
\clearpage
\section{Additional Details on the Realization of MSLs}
Effective time--parallelization of MSLs requires implementation of specialized computational primitives. In example, forward sensitivity methods benefit from a breakdown of matrix--jacobian products into a {\tt vmapped} vector--jacobian products. We have developed a {\tt PyTorch} library designed for broader compatibility with the neural differential equation ecosystem e.g {\tt torchdiffeq} \citep{NEURIPS2018_69386f6b} and {\tt torchdyn} \citep{poli2020torchdyn}. Here, we provide code for several key methods and classes. 

\subsection{Software Implementation of Forward Sensitivity}\label{asec:fsens_soft}
Forward sensitivity analysis is extensively used in MSLs to compute $\dd \phi_{\theta,n}/\dd b_n$ in parallel for each shooting parameter $b_n,~n=0,\dots,N-1$. We showcase how this can be efficiently implemented in {\tt Pytorch} \citep{paszke2019pytorch}. Although the implementation fully accommodates batches $n_b$ of data, i.e. each $b_n$ is a $n_b$ by $n_z$ matrix, we will limit the algorithmic analysis to the unitary batch dimension.  
The forward sensitivity algorithm aims at computing the solution of the differential equation
\[  
    %\overbrace{
    \begin{pmatrix}
        \dot z_n(t)\\
        \dot v_n(t)
    \end{pmatrix} = 
    \begin{pmatrix}
        f_\theta(t, z_n(t))\\
      {\sD f_\theta(t, z_n(t))}v_n(t)
    \end{pmatrix}%
    ,~~~~
    \begin{pmatrix}
        \dot z_n(0)\\
        \dot v_n(0)
    \end{pmatrix} = 
    \begin{pmatrix}
        b_n\\
        \Id_{n_z}
    \end{pmatrix},~~~~
    t\in[t_n,t_{n+1}]={\blue \cT_n}
\]
for all $n$, to return $\phi_{\theta,n}=z_n(t_{n+1})$ and $\dd\phi_{\theta,n}/\dd b_n=v_n(t_{n+1})$. Let $Z$ and $V$ be the tuples containing all $z_n$ and $v_n$,
\[
    \begin{aligned}
        Z &= (z_0, z_1,\dots,z_{N-1})\in \overbrace{\R^{n_z}\times\cdots\times\R^{n_z}}^{N}\equiv\R^{N\times n_z}\\
        V &= (v_0, v_1, \cdots, v_{N-1})\in \underbrace{\R^{n_z\times n_z}\times\cdots\times\R^{n_z\times n_z}}_{N}\equiv\R^{N\times n_z\times n_z}.
    \end{aligned}
\]
Given a tuple of time instants $T=(\tau^0, \tau^1, \dots, \tau^{N-1})\in [t_0,t_1]\times [t_1, t_1]\times [t_{N-1},t_N]\subset \R^{N\times 1}$, $f_\theta$ can evaluated in parallel on $T$ and $Z$ as the number $N$ of shooting parameters $b_n$ and sub--intervals $[t_n, t_{n+1}]$ only accounts for a \textit{batch} dimension. From a software perspective, we can obtain 
\[
    F(T, Z) = (f_\theta(\tau^0, z_0), \dots, f_\theta(\tau^{N-1}, z_{N-1}))
\]
in a single call of the function $f_\theta$, e.g. an instantiated {\tt PyTorch}'s {\tt nn.Module} object. Conversely, when attempting to compute ``$\frac{\partial F}{\partial Z}V$'' in parallel, additional software infrastructure is necessary. The main obstacle is that each Jacobian--matrix product ({\tt jmp})
\[
    \frac{\partial f_\theta (t, z_n(t))}{\partial z_n}v_n(t)
\]
generally requires $n_z$ {\tt autograd} calls. Following the {\tt Jax}'s \citep{jax2018github} approach, we make use of a {\tt PyTorch} implementation\footnote{see \url{https://pytorch.org/docs/master/generated/torch.vmap.html}} of \textit{vectorizing maps} ({\tt vmaps}) to distribute the computation of the individual Jacobian--vectors products (in \textit{batch} for each $n=0,\dots,N-1$) and compose the {\tt jmp} row--by--row or column--by--column. In particular we define the {$\tt vmapped\_jmp$} function
\begin{mintedbox}{python}
    def vmapped_jmp(y, z, v):
        """ Parallel computation of matrix jacobian products with vmap
        """
        def get_jvp(v):
            return torch.autograd.grad(y, z, v, retain_graph=True)[0]
        return vmap(get_jvp, in_dims=2, out_dims=2)(v)
\end{mintedbox}
The forward sensitivity can then be computed as follows
\begin{mintedbox}{python}
class ForwardSensitivity(nn.Module):
    "Forward sensitivity for ODEs. Integrates the ODE returning the state and forward sensitivity"
    def __init__(self, f):
        super().__init__()
        self.f = f

    def forward(self, z0, t_span, method='dopri5', atol=1e-5, rtol=1e-5):
        I = eye(z0.shape[-1]).to(z0)
        # handle regular `batch, dim` case as well as `seq_dim, batch, dim`
        I = I.repeat(z0.shape[0], 1, 1) if len(z0.shape) < 3 else I.repeat(*z0.shape[:2], 1, 1)
        zv0 = (z0, I)
        zT, vT = odeint(self._sensitivity_dynamics, zv0, t_span, method=method, atol=atol, rtol=rtol)
        return zT, vT

    def _sensitivity_dynamics(self, t, zv):
        # z: batch, dim
        # V: batch, dim, dim
        z, v = zv
        # compute vector field
        dz = self.f(t, z.requires_grad_(True))
        # compute fw sensitivity via jmp
        dv = vmapped_jmp(dz, z, v)
        return (dz, dv)
\end{mintedbox}
where {\tt odeint} is ODE solver utility of the {\tt torchdiffeq} \citep{NEURIPS2018_69386f6b}  library.
\subsection{Implementation of Direct Newton Method}
% {\color{blue!70}{\tt fw sensitivity} MSL} as well as {\color{orange!70}zeroth--order MSL}
\textit{Forward sensitivity Newton} ({\color{blue!70}{\tt fw sensitivity}}) MSL is a variant of the proposed model class which obtains the quantities $\sD\phi_{\theta,n}$ directly by augmenting the time--parallelized forward dynamics through the {\tt ForwardSensitivity} class previously detailed. During the evaluation of the \textit{advancement} function $\gamma_\theta(B, z_0)=(z_0, \phi_\theta(b_0, t_0, t_1),\dots,\phi_\theta(b_{N-1}, t_{N-1}, t_N))$, {\tt ForwardSensitivity} maximizes reutilization of vector field $f_\theta$ evaluations by leveraging the results to advance both \textit{standard} as well as sensitivity dynamics. This provides an overall reduction in the potentially expensive evaluation of the neural network $f_\theta$, compared to \textit{parareal} ({\color{orange!70}zeroth--order MSL}). We hereby report the {\tt PyTorch} implementation for both the    {\color{blue!70}{\tt fw sensitivity} MSL} and {\color{orange!70}zeroth--order MSL} methods
\begin{mintedbox}{python}
def functional_msdirect_forward(vf, z, B, fsens, maxiter, sub_t_span, n_sub, fine_method, coarse_method, fine_rtol, fine_atol):
        """Forward pass of Multiple Shooting Layer (MSL) layer via parallelized forward sensitivity analysis on the flows to compute the Jabians in the correction term of the Newton iteration"""
    i = 0
    while i <= maxiter:
        i += 1
        with torch.set_grad_enabled(True):
            B_fine, V_fine = fsens(B[i - 1:], sub_t_span, method=fine_method, rtol=fine_rtol, atol=fine_atol)
        B_fine, V_fine = B_fine[-1], V_fine[-1]

        B_out = torch.zeros_like(B)
        B_out[:i] = B[:i]
        B_in = B[i - 1]
        for m in range(i, n_sub):
            B_in = B_fine[m - i] + torch.einsum('bij, bj -> bi', V_fine[m - i], B_in - B[m - 1])
            B_out[m] = B_in
        B = B_out
    return B
\end{mintedbox}
\begin{mintedbox}{python}
def functional_mszero_forward(vf, z, B, fsens, maxiter, sub_t_span, n_sub, fine_method, coarse_method, fine_rtol, fine_atol):
    """Forward pass of Multiple Shooting Layer (MSL) layer via zeroth-order (parareal) approximation of the correction term"""
    i = 0
    while i <= maxiter:
        i += 1
        B_coarse = odeint(vf, B[i - 1:], sub_t_span, method=coarse_method)[-1]
        B_fine = odeint(vf, B[i - 1:], sub_t_span, method=fine_method, rtol=fine_rtol, atol=fine_atol)[-1]

        B_out = torch.zeros_like(B)
        B_out[:i] = B[:i]
        B_in = B[i - 1]
        for m in range(i, n_sub):
            B_in = odeint(vf, B_in, sub_t_span, method=coarse_method)[-1]
            B_in = B_in - B_coarse[m - i] + B_fine[m - i]
            B_out[m] = B_in
        B = B_out
    return B
\end{mintedbox}
In the above, we employ the finite--step convergence property of Newton MSL iterations to avoid redundant computation. More specifically, at iteration $k$ we do not advance shooting parameters $b_n,~~n < k$ by slicing the tensor $B$ during $\gamma_\theta$ evaluations. Similarly, updates in the form \eqref{eq:7} are not performed for shooting parameters already at convergence.
\subsection{Alternative Approaches to MSL Inference}
\paragraph{On Newton and Quasi-Newton methods for MSL}
The root--finding problem arising in MSLs can also be approached by standard application of Newton or Quasi--Newton algorithms. Although Quasi--Newton algorithms can provide  improved computational efficiency by maintaining a low--rank approximation of the Jacobian $Dg_\theta(B, z_0)$ rather than computing it from scratch every iteration, this advantage does not translate well to the MSL case. Popular examples include, e.g., the Broyden family \cite{broyden1965class} employed in \textit{Deep Equilibrium Models} (DEQs) \cite{bai2019deep}. As discussed in the main text, thanks to the special structure of the Jacobian of the MSL problem, the direct Newton algorithm \eqref{eq:7} can be applied without computation and inversion of the full Jacobian.
\paragraph{Root finding via gradient descent}
% %
A completely different approach to solve the implicit forward MSL pass \eqref{eq:3} is to tackle the root--finding via some gradient--descent (GD) method minimizing $\|g_{\theta}(B)\|_2^2$, i.e.
\[
    B^* = \argmin \frac{1}{2}\|g_\theta(B)\|_2^2.
\]
In the case of MSL, all GD solutions (i.e. minima of $\|g_\theta(B)\|_2^2$) are the same of the the root finding ones. This can be intuitively checked by inspecting the zeros of the gradient, i.e.
\[
    \nabla_B\frac{1}{2}\|g_\theta(B)\|_2^2 = Dg_\theta|_{B}g_\theta(B)
\]
and, since $Dg_\theta|_{B}$ is nonsingular for all $B$,  
\[
    \forall \tilde{B}^* ~:~ \nabla_B\frac{1}{2}\|g_\theta(\tilde{B}^*)\|_2^2 = \0 \Rightarrow g_\theta(\tilde{B}^*)=\0.
\]
\subsection{Implementation of Backward Interpolated Adjoint}\label{asec:backprop_impl}
We provide pseudo--code for our implementation of MSLs with backward gradients obtained via interpolated adjoints. The implementation relies on cubic interpolation utilities provided by {\tt torchcde} \cite{kidger2020neural}. Interpolation is used to obtain values of $z(t)$ without a full backsolve from $z(T)$.
\begin{mintedbox}[fontfamily=tt]{python}
def msfunc_interp_adjoint(vf, f_params, fsens, t_span, sub_t_span, fine_method, fine_atol, fine_rtol, coarse_method, coarse_atol, coarse_rtol, num_subintervals, maxiters, func_forward):
    """
        Return Multiple Shooting Layer with backward gradients computed via interpolated adjoints. 
    """
    class MSFunction(Function):
        @staticmethod
        def forward(ctx, f_params, z, B):
            B = FORWARD_TYPE_MAP[func_forward](vf, z, B, fsens, maxiters, sub_t_span, num_subintervals, fine_method, coarse_method, fine_rtol, fine_atol)
            ctx.save_for_backward(B)
            return B

        @staticmethod
        def backward(ctx, *grad_output):
            z,  = ctx.saved_tensors
            # create interpolation
            spline_coeffs = natural_cubic_spline_coeffs(t_span.to(z), z.permute(1, 0, 2).detach())
            z_spline = NaturalCubicSpline(t_span, spline_coeffs)
            
            # define adjoint dynamics
            def adjoint_dynamics(s, adj):
                λ, μ = adj[0:2]
                with torch.set_grad_enabled(True):
                    z = z_spline.evaluate(s).requires_grad_(True)
                    dzds = vf(s, z)
                    dλds = grad(dzds, z, -λ, allow_unused=True, retain_graph=True)[0]
                    dμds = tuple(grad(dzds, tuple(vf.parameters()), -λ, allow_unused=True, retain_graph=False))
                    # Safety checks for `None` gradients
                    dμds = torch.cat([el.flatten() if el is not None else torch.zeros_like(p) for el, p in zip(dμds, vf.parameters())])
                return (dλds, dμds)

            # init adjoint state
            # λ0 dimensions: `batch, batch_seq, dim`
            # λ0 dimensions: tuple(`param_dim`) for each param group
            λ0 = grad_output[0][-1]
            f_params = torch.cat([p.contiguous().flatten() for p in vf.parameters()])
            μ0 = torch.zeros_like(f_params)
            adjoint_state = (λ0, μ0)

            # solve the adjoint
            for i in range(len(t_span) - 1, 0, -1):
                adj_sol = torchdiffeq.odeint(adjoint_dynamics, adjoint_state, t_span[i-1:i+1].flip(0), rtol=1e-7, atol=1e-7, method='dopri5')

                # prepare adjoint state for next interval
                #print(grad_flipped[i][0])
                λ = adj_sol[0][-1] + grad_output[0][i-1]
                μ = adj_sol[1][-1]
                adjoint_state = (λ, μ)
            return (μ, λ, None)

    return MSFunction
\end{mintedbox}

\subsection{Broader Impact}
Differential equations are the language of science and engineering. As methods \citep{jia2019neural} and software frameworks \citep{rackauckas2019diffeqflux,li2020scalable,poli2020torchdyn} are improved, yielding performance gains or speedups \citep{poli2020hypersolvers,kidger2020hey,pal2021opening}, the range of applicability of neural differential equations is extended to more complex and larger scale problems. As with other techniques designed to reduce overall training time, we expect a net positive environment impact from the adoption of MSLs in the framework.

Application domains for MSLs include environments with real--time constraints, for example control and high frequency time series prediction. Shorter inference wall--clock and training iteration times should yield more robust models that can, in example, be retrained online at higher frequencies as more data is collected.

%% file: Appendix/Appendix_D.tex
\clearpage
\section{Neural Network Control of the Timoshenko Beam}\label{asec:pde}
In this section we derive the dynamic model of the Timoshenko beam, the boundary control and the structure--preserving discretization of the problem.
\subsection{Port--Based Modeling of the Timoshenko Beam}
Linear distributed port-Hamiltonian systems \citep{macchelli2004port} in one-dimensional domains take the form
\begin{equation}\label{eq:pHsys}
    \frac{\partial z}{\partial t}(x,t) = P_1 \frac{\partial }{\partial x}(\mathcal{L}(x)z(x,t)) + (P_0-G_0)\mathcal{L}(x)z(x,t)
\end{equation}
with distributed state $z \in \R^{n_z}$ and spatial variable $x \in [a, b]$. Moreover, $P_1 = P_1^\top$ and
invertible, $P_0 = -P_0^\top,\; G_0 = G_0^\top \ge 0$, and $\mathcal{L}(\cdot)$ is a bounded and Lipschitz continuous matrix-valued function such that $\mathcal{L}(x) = \mathcal{L}^\top(x)$ and $\mathcal{L}(x) \ge \kappa I$, with $\kappa > 0$, $\forall x \in [a, b]$.
Given the Hamiltonian (total energy) of the system
$$H= \norm{z}^2_{\mathcal{L}} = \inner[L^2]{z}{\mathcal{L}z},$$
its variational derivative corresponds to the term $\mathcal{L}(x)z(x,t)$:
\begin{equation*}
    \frac{\delta H}{\delta z}(z(x,t), x) = \mathcal{L}(x) z(x,t) 
\end{equation*}

A particular example from continuum mechanics that falls within the systems class \eqref{eq:pHsys} is the Timoshenko beam with no dissipation \citep{macchelli2004modeling}. This system takes the following form:
\begin{equation}\label{eq:pHtimo}
    \frac{\partial}{\partial t}
    \begin{pmatrix}
    p_t \\ p_r \\ \varepsilon_r \\ \varepsilon_t
    \end{pmatrix} = 
    \begin{bmatrix}
    0 & 0 & 0 & \partial_z \\
    0 & 0 & \partial_z & 1 \\
    0 & \partial_z & 0 & 0 \\
    \partial_z & -1 & 0 & 0
    \end{bmatrix}
    \left(
    \begin{bmatrix}
    (\rho A)^{-1} & 0 & 0 & 0 \\
    0 & (I_\rho)^{-1} & 0 & 0 \\
    0 & 0 & EI & 0 \\
    0 & 0 & 0 & K_{\mathrm{sh}} G A
    \end{bmatrix}
    \begin{pmatrix}
    p_t \\ p_r \\ \varepsilon_r \\ \varepsilon_t
    \end{pmatrix}
    \right),
\end{equation}
where $\rho$ is the mass density, $A$ is the cross section area, $I_\rho$ is the rotational inertia, $E$ is the Young modulus, $I$ the cross section moment of area, $K_{\mathrm{sh}}=5/6$ is the shear correction factor and $G$ the shear modulus. \\

For this examples the matrices $P_0, \; G_0, \; P_1, \; \mathcal{L}$ and  are given by
\begin{equation}\label{eq:mat_timo}
    \begin{aligned}
        P_0 = 
        \begin{bmatrix}
        0 & 0 & 0 & 0 \\
        0 & 0 & 0 & 1 \\
        0 & 0 & 0 & 0 \\
        0 & -1 & 0 & 0
        \end{bmatrix}, \quad
        P_1 = 
        \begin{bmatrix}
        0 & 0 & 0 & 1 \\
        0 & 0 & 1 & 0 \\
        0 & 1 & 0 & 0 \\
        1 & 0 & 0 & 0
        \end{bmatrix},  \quad
        G_0 = \0_{4 \times 4},\\
        \mathcal{L}(z) =
        \begin{bmatrix}
            (\rho A)^{-1} & 0 & 0 & 0 \\
            0 & (I_\rho)^{-1} & 0 & 0 \\
            0 & 0 & EI & 0 \\
            0 & 0 & 0 & K_{\mathrm{sh}} G A
        \end{bmatrix}.
    \end{aligned}
\end{equation}

We investigate the boundary control of the Timoshenko beam model. As control input, the following selection is made (cantilever-free beam)
\begin{equation}
\pi_\partial = 
\begin{pmatrix}
EI \varepsilon_r(b, t)\\
K_{\mathrm{sh}}GA \varepsilon_t(b,t)\\
(\rho A)^{-1} p_t(a,t)\\
(I_\rho)^{-1} p_r(a,t)\\
\end{pmatrix}
\end{equation}
Notice that the control expression can be rewritten compactly as follows

\begin{equation}\label{eq:u_partial}
\pi_\partial = 
\mathcal{B}_\partial
\begin{pmatrix}
\mathcal{L}x(b, t)\\
\mathcal{L}x(a, t)\\
\end{pmatrix}, \quad\text{where}\quad
\mathcal{B}_\partial  = \begin{bmatrix}
0 & 0 & 1 & 0 & 0 & 0 & 0 & 0\\ 
0 & 0 & 0 & 1 & 0 & 0 & 0 & 0\\ 
0 & 0 & 0 & 0 & 1 & 0 & 0 & 0\\ 
0 & 0 & 0 & 0 & 0 & 1 & 0 & 0\\ 
\end{bmatrix}.
\end{equation}
To put system \eqref{eq:pHtimo} in impedance form, the outputs are selected as follows

\begin{equation}
y_{\partial} =
\begin{pmatrix}
(\rho A)^{-1} p_t(b,t)\\
(I_\rho)^{-1} p_r(b,t)\\
-EI \varepsilon_r(a, t)\\
-K_{\mathrm{sh}}GA \varepsilon_t(a,t)\\
\end{pmatrix}
\end{equation}
This is compactly written as 
\begin{equation}\label{eq:y_partial}
y_\partial = 
\mathcal{C}_\partial
\begin{pmatrix}
\mathcal{L}x(b, t)\\
\mathcal{L}x(a, t)\\
\end{pmatrix}, \quad\text{where}\quad
\mathcal{C}_\partial  = \begin{bmatrix}
1 & 0 & 0 & 0 & 0 & 0 & 0 & 0\\ 
0 & 1 & 0 & 0 & 0 & 0 & 0 & 0\\ 
0 & 0 & 0 & 0 & 0 & 0 & -1 & 0\\ 
0 & 0 & 0 & 0 & 0 & 0 & 0 & -1\\ 
\end{bmatrix}.
\end{equation}

With this selection of inputs and outputs, the rate of the Hamiltonian  is readily computed
\begin{equation}
    \dot{H} = \pi_{\partial}^\top y_{\partial}.
\end{equation}
Within the purpose of this paper we restrict to the case of \textit{a cantilever beam undergoing a control action at the free end}
\begin{equation}
    \pi_\partial =
    \begin{pmatrix}
    \pi_{\partial, 1} \\ \pi_{\partial, 2} \\ 0 \\ 0 \\
    \end{pmatrix},
\end{equation}
where $\pi_{\partial, 1}$ is the control torque and $u_{\partial, 2}$ is the control force.

\subsection{Discretization of the Problem}
To discretize system \eqref{eq:pHtimo}, since the problem is linear, one can either rely on a energy formulation or a co-energy one. Given the coenergy variables
\begin{equation}
    \begin{pmatrix}
    v_t \\ v_r \\ \sigma_r \\ \sigma_t \\
    \end{pmatrix} =
    \begin{bmatrix}
    (\rho A)^{-1} & 0 & 0 & 0 \\
    0 & I_\rho^{-1} & 0 & 0 \\
    0 & 0 & EI & 0 \\
    0 & 0 & 0 & K_{\text{sh}} G A
    \end{bmatrix}
    \begin{pmatrix}
    p_t \\ p_r \\ \varepsilon_r \\ \varepsilon_t
    \end{pmatrix}
\end{equation}
and introducing the bending and shear compliance
\begin{equation}
    C_b = (EI)^{-1}, \qquad C_s = (K_{\text{sh}} G A)^{-1},
\end{equation}
system \eqref{eq:pHtimo} is rewritten as

\begin{equation}\label{eq:pHtimo_coenergy}
    \begin{bmatrix}
    \rho A & 0 & 0 & 0 \\
    0 & I_\rho & 0 & 0 \\
    0 & 0 & C_b & 0 \\
    0 & 0 & 0 & C_s
    \end{bmatrix}
    \frac{\partial}{\partial t}
    \begin{pmatrix}
    v_t \\ v_r \\ \sigma_r \\ \sigma_t \\
    \end{pmatrix} = 
    \begin{bmatrix}
    0 & 0 & 0 & \partial_z \\
    0 & 0 & \partial_z & 1 \\
    0 & \partial_z & 0 & 0 \\
    \partial_z & -1 & 0 & 0
    \end{bmatrix}
    \begin{pmatrix}
    v_t \\ v_r \\ \sigma_r \\ \sigma_t \\
    \end{pmatrix},
\end{equation}

A weak form suitable for mixed finite elements is readily obtained by considering its weak form using test functions $(\mu_t, \mu_r, \nu_r, \nu_t)$ and the integration by parts applied to the first two lines. In this formulation, the Dirichlet boundary condition have to be incorporated as essential boundary conditions
\begin{equation}
    \begin{aligned}
    \inner[\Omega]{\mu_t}{\rho A \partial_t v_t} &= - \inner[\Omega]{\partial_z\mu_t}{\sigma_t} + \mu_t(b) \pi_{\partial, 2}, \\  
    \inner[\Omega]{\mu_r}{I_\rho \partial_t v_r} &= -\inner[\Omega]{\partial_z\mu_r}{\sigma_r} + \inner[\Omega]{\mu_r}{\sigma_t} + \mu_r(b) \pi_{\partial, 1}, \\
    \inner[\Omega]{\nu_r}{C_b \partial_t \sigma_r} &= \inner[\Omega]{\nu_r}{\partial_z v_r}, \\  
    \inner[\Omega]{\nu_t}{C_s \partial_t \sigma_t} &= \inner[\Omega]{\nu_t}{\partial_z v_t} - \inner[\Omega]{\nu_t}{v_r},
    \end{aligned}
\end{equation}

where $\Omega =[a, b]$ and $\inner[\Omega]{f}{g} = \int_a^b f g \dd{x}$. Introducing the following Galerkin basis functions

\begin{equation}\label{eq:approx_timo}
\begin{aligned}
{\mu}_t = \sum_{i = 1}^{N_{v_t}} \varphi_{v_t}^i \mu_t^i, \qquad
{\mu}_r = \sum_{i = 1}^{N_{v_r}} \varphi_{v_r}^i \mu_r^i, \qquad
{\nu}_r = \sum_{i = 1}^{N_{\sigma_r}} \varphi_{\sigma_r}^i \nu_r^i, \qquad
{\nu}_t = \sum_{i = 1}^{N_{\sigma_t}} \varphi_{\sigma_t}^i \nu_r^i,  \\
{v}_t = \sum_{i = 1}^{N_{v_t}} \varphi_{v_t}^i v_t^i, \qquad
{v}_r = \sum_{i = 1}^{N_{v_r}} \varphi_{v_r}^i v_r^i, \qquad
{\sigma}_r = \sum_{i = 1}^{N_{\sigma_r}} \varphi_{\sigma_r}^i \sigma_r^i, \qquad
{\sigma}_t = \sum_{i = 1}^{N_{\sigma_t}} \varphi_{\sigma_t}^i \sigma_r^i,
\end{aligned}
\end{equation}  
a finite-dimensional system is obtained

\begin{equation}\label{eq:pHfindim_timo_grad}
\begin{aligned}
\begin{bmatrix}
{M}_{\rho A} & \x & \x & \x \\
\x & {M}_{I_\rho}& \x & \x \\
\x & \x & {M}_{C_b} & \x \\
\x & \x & \x & {M}_{C_s}\\
\end{bmatrix}
\begin{bmatrix}
\ubar{\dot{{v}}}_t \\
\ubar{\dot{{v}}}_r \\
\ubar{\dot{{\sigma}}}_r \\
\ubar{\dot{{\sigma}}}_t \\
\end{bmatrix}
&= \begin{bmatrix}
\x &  \x &  \x &   -{D}_{1}^\top \\
\x &  \x &  -{D}_{2}^\top & -{D}_{0}^\top \\
\x &  {D}_{2} & \x & \x\\
{D}_{1} & {D}_{0} & \x & \x\\
\end{bmatrix} 
\begin{bmatrix}
\ubar{{v}}_t \\
\ubar{{v}}_r \\
\ubar{{\sigma}}_r \\
\ubar{{\sigma}}_t \\
\end{bmatrix} + 
\begin{bmatrix}
\x & {B}_F \\
{B}_{T} & \x\\
\x & \x \\
\x & \x  \\
\end{bmatrix}
\begin{bmatrix}
{\pi}_{\partial, 1} \\
{\pi}_{\partial, 2}
\end{bmatrix}, \\
\begin{bmatrix}
{y}_{\partial, 1} \\
{y}_{\partial, 2}
\end{bmatrix} &= \begin{bmatrix}
\x & {B}_T & \x & \x \\
{B}_F & \x & \x & \x \\
\end{bmatrix}
\begin{bmatrix}
\ubar{{v}}_t \\
\ubar{{v}}_r \\
\ubar{{\sigma}}_r \\
\ubar{{\sigma}}_t \\
\end{bmatrix}.
\end{aligned}
\end{equation}

The mass matrices ${M}_{\rho h}, \; {M}_{I_\theta}, \; {M}_{\bm{\mathcal{C}}_b}, \; {M}_{{C}_s}$ are computed as
\begin{equation}
\begin{aligned}
M_{\rho A}^{ij} &= \inner[\Omega]{\varphi_{v_t}^i}{\rho A \varphi_{v_t}^j}, \\
M_{I_\rho}^{mn} &= \inner[\Omega]{\varphi_{v_r}^m}{I_\rho \varphi_{v_r}^n},
\end{aligned} \qquad
\begin{aligned}
\quad M_{C_b}^{pq} &= \inner[\Omega]{\varphi_{\sigma_r}^p}{C_b \varphi_{\sigma_r}^q}, \\
M_{C_s}^{rs} &= \inner[\Omega]{\varphi_{\sigma_t}^l}{C_s\varphi_{\sigma_t}^s}, 
\end{aligned}
\end{equation}
where $i, j \in \{1, N_{v_t}\}, \; m, n \in \{1, N_{v_r}\}, \, p, q \in \{1, N_{\sigma_r}\}, \; l, s \in \{1, N_{\sigma_t}\}$. Matrices ${D}_{1}, \; {D}_{2}, \; {D}_{0}$ assume the form
\begin{equation}
\begin{aligned}
D_{1}^{lj} &= \inner[\Omega]{\varphi_{\sigma_t}^l}{\partial_z \varphi_{v_T}^j}, \\ D_{2}^{pn} &= \inner[\Omega]{\varphi_{\sigma_r}^p}{\partial_z \varphi_{v_r}^n},
\end{aligned} \qquad
D_{0}^{rn} = -\inner[\Omega]{\varphi_{\sigma_t}^r}{\varphi_{v_r}^n}.
\end{equation}

Vectors ${B}_F, \, {B}_T$ are computed as ($i \in {1,N_{v_t}}$ and ($m \in {1,N_{v_r}}$)
\begin{equation}
{B}_F^i = \varphi_{v_t}^i(b), \qquad {B}_{T}^j = \varphi_{v_r}^j(b).
\end{equation}
\subsection{Control by Neural Approximators and MSL}
Due to invertibility of the mass matrix we can reduce the above equation to a controlled linear system representing the discretized dynamics of the boudary-controlled Tymoshenko beam 
\begin{equation}\label{aeq:fe_model}
    \begin{aligned}
        \ubar{\dot z}(t) &= A \ubar z(t) + B \pi_\partial(t)\\
        y_\partial(t) &= C \ubar z(t)
    \end{aligned}
\end{equation}
with
\begin{equation}
    z = \begin{bmatrix}
\ubar{{v}}_t \\
\ubar{{v}}_r \\
\ubar{{\sigma}}_r \\
\ubar{{\sigma}}_t \\
\end{bmatrix},\quad \pi_\partial = \begin{bmatrix}
{\pi}_{\partial, 1} \\
{\pi}_{\partial, 2}
\end{bmatrix}, \quad y_\partial = 
\begin{bmatrix}
{y}_{\partial, 1} \\
{y}_{\partial, 2}
\end{bmatrix}
\end{equation}
and 
\begin{equation}
    \begin{aligned}
        A =  
        \begin{bmatrix}
            \x &  \x &  \x &   -{M}_{\rho A}^{-1}{D}_{1}^\top \\
            \x &  \x &  -{M}_{I_\rho}^{-1}{D}_{2}^\top & -{M}_{I_\rho}^{-1}{D}_{0}^\top \\
            \x &  {M}_{C_b}^{-1}{D}_{2} & \x & \x\\
            {M}_{C_s}^{-1}{D}_{1} & {M}_{C_s}^{-1}{D}_{0} & \x & \x\\
        \end{bmatrix},\\ 
        B = 
        \begin{bmatrix}
            \x & {M}_{\rho A}^{-1}{B}_F\\
            {M}_{I_\rho}^{-1}{B}_T & \x &\\
            \x & \x \\
            \x & \x \\
        \end{bmatrix},\quad 
        C = 
        \begin{bmatrix}
            \x & {B}_T & \x & \x \\
            {B}_F & \x & \x & \x \\
        \end{bmatrix}
    \end{aligned}
\end{equation}
We consider a parametrization $u_{\partial, \theta}$ with parameters $\theta$ of the boundary controller $\pi_\partial$ via a multi--layer perceptron. The neural network controller $\pi_{\partial, \theta}$ takes as input the discretized state of the PDE $\pi_\partial(t) = \pi_{\partial,\theta}(\ubar z(t)),~~t\mapsto z\mapsto u_{\partial,\theta}$. We apply the MSL to the controlled system 
\[
    \ubar{\dot z}(t) = A\ubar z(t) + B\pi_{\partial,\theta}(\ubar z(t))
\]
Further details on the experimental setup and numerical results are given in Appendix \ref{asec:exp_pde}.

%% file: Appendix/Appendix_E.tex
\clearpage
\section{Experimental Details}
\paragraph{Experimental setup} Experiments have been performed on a workstation equipped with a 48 threads \textsc{AMD Ryzen Threadripper 3960X} a \textsc{NVIDIA GeForce RTX 3090} GPUs and two \textsc{NVIDIA RTX A6000}. The main software implementation has been done within the $\tt PyTorch$ framework. Some functionalities rely on {\tt torchdiffeq} \citep{NEURIPS2018_69386f6b} ODE solvers and {\tt torchcde} \citep{kidger2020neural} cubic splines interpolation utilities for the \textit{interpolated} version of the adjoint gradients. 
\paragraph{Common experimental settings} In all experiments to setup the multiple shooting problem, we choose an evenly spaced discretization of the time domain $[t_0,t_N]$, i.e.
\[
    \forall n=1,\dots,N~~~~t_n = t_{n-1} + \frac{1}{N}(t_N-t_0)
\]
\subsection{Variational Multiple Shooting Layers}\label{asec:exp_vml}
\paragraph{Dataset}
We apply \textit{variational multiple shooting layers} (vMSL) to trajectory generations of various dynamical systems. In particular, we consider the Van Der Pol oscillator
\begin{equation*}
    \begin{aligned}
        \dot p &= q \\
        \dot q &= \alpha(1 - p^2)q - p
    \end{aligned}
\end{equation*}
as well as the Rayleigh Duffing system
\begin{equation*}
    \begin{aligned}
        \dot p &= q \\
        \dot q &= \alpha p - 2 p^3 + (1 - q^2)q
    \end{aligned}
\end{equation*}
We generate a dataset of $10000$ trajectories by solving the above systems until $T=1$. Each trajectory consists of $20$ regularly sampled observations subject to additive noise $\epsilon$ where $\epsilon\sim\mathcal{N}(0, \Sigma)$, with $\Sigma$ not diagonal i.e state--correlated noise.
\paragraph{Models and training}
Both vMSLs as well as Latent Neural ODE baselines are trained for 
Latent Neural GDEs are trained for $300$ epochs with {\textsc{Adam}} \citep{kingma2014adam}. We schedule the learning rate using one cycle policies  \citep{smith2019super} where the cycle peak for the learning rate is $10^{-2}$, set to be reached at epoch $100$. The encoder architecture is shared across all models as is defined as two layers of \textit{temporal convolutions} (TCNs), followed by a linear layer operating on flattened features. Between each TCN layer we introduce a maxpool operator to reduce sequence length. We solve Neural ODEs with {\tt dopri5} solver with tolerances $10^{-4}$. 

\begin{figure}
    \centering
    \includegraphics[width=\linewidth]{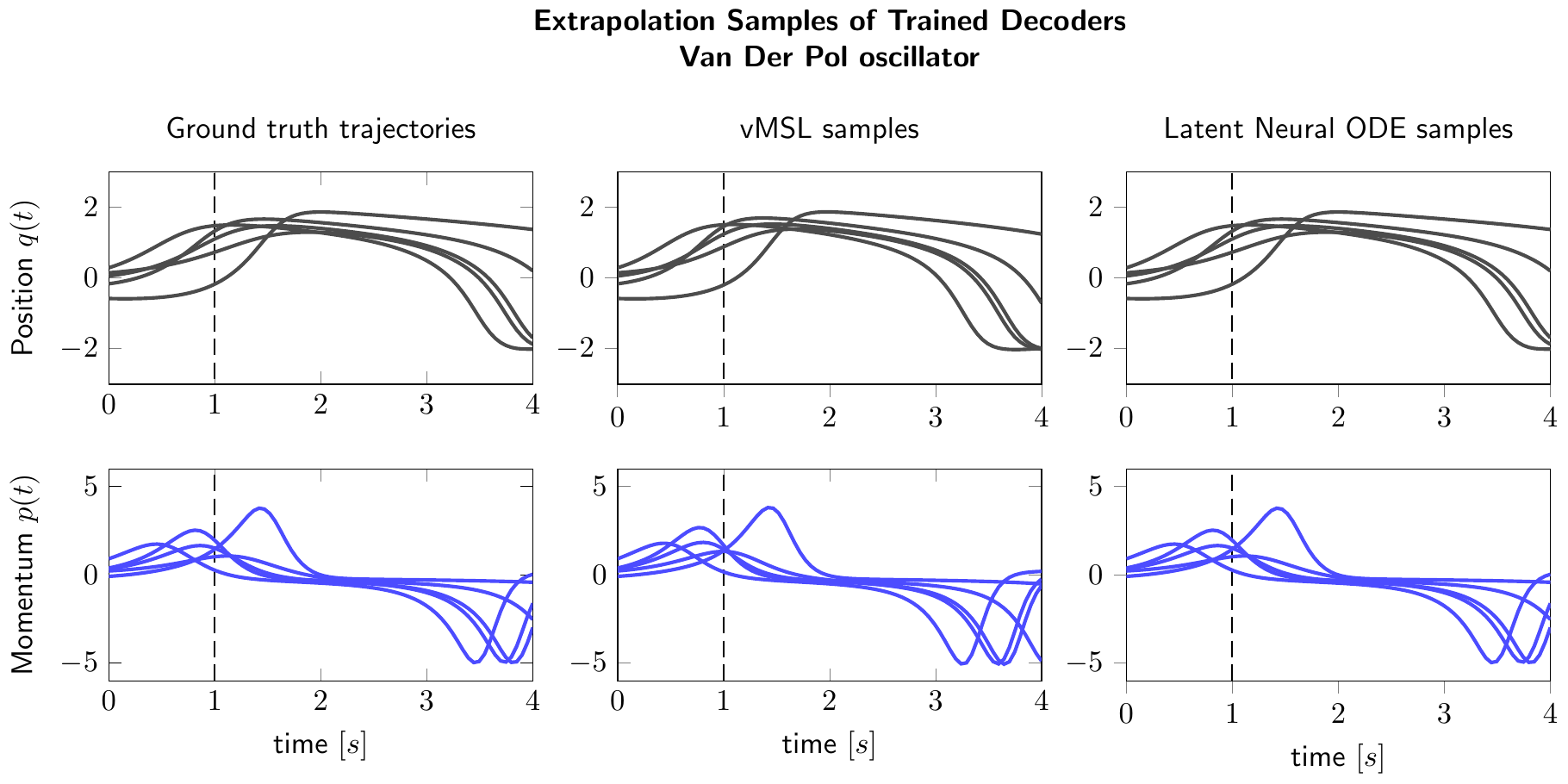}
    \caption{\footnotesize Samples of vMSLs and Latent Neural ODE baselines in the trajectory generation task on Van Der Pol oscillators. The samples are obtained by querying the decoders at desired initial conditions. The models extrapolate beyond $T=1$ used in training.}
    \label{fig:vmsl_samples}
\end{figure}

\begin{figure}
    \centering
    \includegraphics{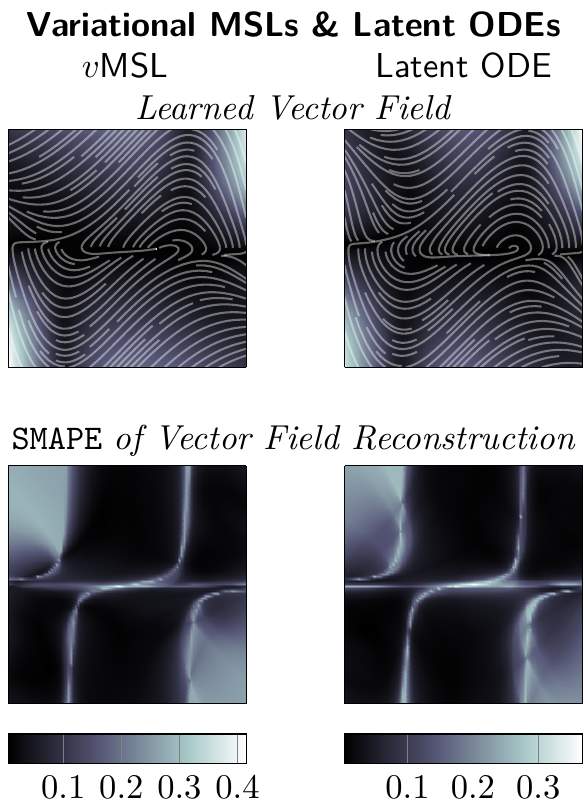}
    \caption{\footnotesize Learned vector fields by vMSL and Latent ODE decoders trained on noisy trajectories of the Van der Pol oscillator. vMSL models obtain the same result at a significantly cheaper NFE cost.}
    \label{fig:vmsl_vf}
\end{figure}

We experiment with both {\color{blue!70}{\tt fw sensitivity} MSL} as well as {\color{orange!70}zeroth--order MSL} as vMSL decoders. In all cases, we perform a single iteration of the chosen forward method. The parallelized ODE solves apply a single step of \textit{Runge--Kutta 4}. We note that vMSL \textit{number of function evaluation} (NFE) measurements also include the initialization calls to the vector field performed by the coarse solver to obtain shooting parameters $B_0$. Fig.~\ref{fig:vmsl_samples} provides visualizations for decoder samples (extrapolation) of all models compared to ground--truth trajectories while Fig.~\ref{fig:vmsl_vf} displays the learned vector fields of both vMSL and Latent ODE model.

To train all models we set the output--space prior $p(\hat{x}):=\mathcal{N}(x, \sigma)$ with $\sigma=0.1$.
\subsection{Optimal Limit Cycle Control via Multiple Shooting Layers}\label{asec:exp_ctrl}
\begin{figure}[h]
    \centering
    \vspace{-5mm}
    \includegraphics{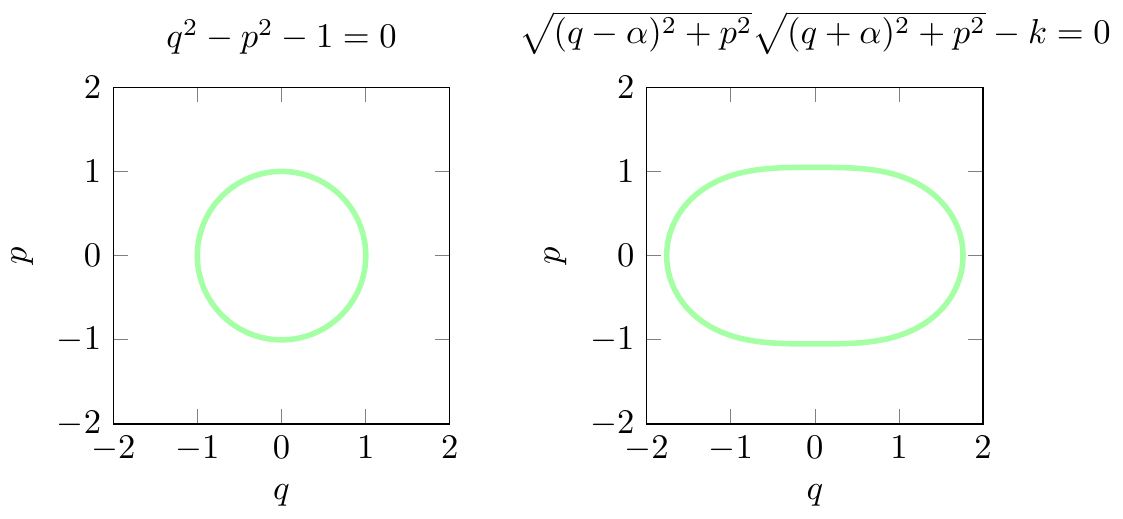}
    \vspace{-3mm}
    \caption{\footnotesize Desired \textit{loci} in the state space, i.e. \textit{limit cycles} to be created in the vector field through the control action $u_\theta(q(t), p(t))$.}
    \label{fig:limit_cycles}
\end{figure}

In the optimal control tasks we considered a simple mechanical system of the form 
\[
    \begin{aligned}
        \dot q(t) &= p(t)\\
        \dot p(t) &= \pi_\theta(q(t),p(t))        
    \end{aligned},~~~~z=[q,p].
\]
evolving in a time span $[t_0, t_N] = [0,10]$ and we fixed $N=99$. The task was the one of stabilizing the state of different \textit{loci} $S_d=\{z\in\cZ:s_d(z)=0\}$ by minimizing $|s_d(z(t))|$, $s_d:[q(t),p(t)]\mapsto s_d(q(t),p(t))$. Specifically, we chose the following \textit{loci} of points
\[  
    \begin{aligned}
        1.~~~~~~~~& s_d(q(t), p(t)) = q^2(t) + p^2(t) - 1~~~~[\text{\blue unit circle}]\\
        2.~~~~~~~~& s_d(q(t), p(t)) = \sqrt{(q(t) - \alpha)^2 + p^2(t)}\sqrt{(q(t) + \alpha)^2 + p^2(t)} - k
    \end{aligned}
\]
across timestamps. The desired curves $s_d$ are displayed in Fig.~\ref{fig:limit_cycles}.

We compared the performance of MSL with the one of a standard (sequential) Neural ODE trained with {\tt dopri5} and {\tt rk4} solver. The objective was to show that MSL can achieve the same control performance while drastically reducing the computational cost of the training.
\paragraph{Models and training}
{
The loss function used to train the controlled was chosen as
$$
    \begin{aligned}
        \min_{\theta}~~~&\frac{1}{2N|Z_0|}\sum_{j=0}^{|Z_0|}\sum_{n=0}^N\left|s_d\left(b_{n,j}^*\right)\right| + \alpha\left\|\pi_\theta(b^*_{n,j})\right\|_1\\
    \text{subject to}~~&B^*_j : g_\theta(B^*_j, z_0^j)=\0\\
    & z_0^j\in Z_0,~~~\alpha \ge 0
    \end{aligned}
$$
where $B^*_j = (b_{0,j}^*~\cdots~b_{N,j}^*)$. It penalizes the distance of trajectories from the desired curve as well as the control effort.} In both the MSL and the Neural ODE baseline the controller $\pi_\theta(q, p)$ has been chosen as a neural network composed with two fully--connected layers of 32 neurons each and hyperbolic tangent activation. 
In the forward pass of MSL we performed a single iteration of the {\color{blue!70}{\tt fw sensitivity}}--type algorithm. The parallelized ODE solver applies a single step of \textit{Runge--Kutta 4} to each shooting parameter $b_n$. The backward pass has been instead performed with reverse--mode AD. At the beginning of the training phase, the shooting parameters $B^0_0$ have been initialized with with the sequential {\tt dopri5} solver with tolerances set to $10^{-8}$, i.e. $B^0_0 = \{\tilde\phi_\theta(z_0, t_0, t_n)\}_n$. As described in the main text, $B^0$ has then been updated at each optimization step with the $B^*$ of the previous iteration to track the changes in the parameters $\theta$ and preserving the ability to track the ``true'' solution $\{\tilde\phi_\theta(z_0, t_0, t_n)\}_n$ with a single iteration of the Newton method (following the results of Theorem \ref{th:track}). The time horizon has been set to $[0, 10s]$ and we fixed $N=100$ shooting parameters. The baseline Neural ODE has been instead trained with standard {\tt dopri5} solver with tolerances set to $10^{-5}$ and the sequential {\tt rk4} solver with $N$ steps over the time horizon.

It is worth to be noticed that both the \textit{parallelized} {\tt rk4} integration step of MSL and the \textit{sequential} {\tt rk4} integration in the Neural ODE baseline operates with the same step size of $0.1s$.

All models have been trained for 2500 epochs with a single batch of 2048 initial conditions $(q_0,p_0)$ uniformly distributed in $[-2,2]\times[-2,2]$ with {\textsc{Adam}} \citep{kingma2014adam} optimizer and learning rate $10^{-4}$. 

For the \textit{circle} desired limit cycle, the training procedure has been repeated with different initial conditions and neural network initializations in a Monte Carlo Simulation of 50 runs. Further, at each training step of MSL we solved the forward system using {\tt dopri5} with absolute and relative tolerances set to $10^{-5}$ to compute {\tt SMAPE} with the current MSL solution across training iterations shown in Fig.~\ref{fig:cmsl_nfe}. Similarly, throughout the training of the baseline Neural ODE we recorded the {\tt NFE}s of the forward pass across iterations.
We also repeated the training of each model recording the wall--clock time of every training iteration.

\begin{figure}
    \centering
    \includegraphics[width=\linewidth]{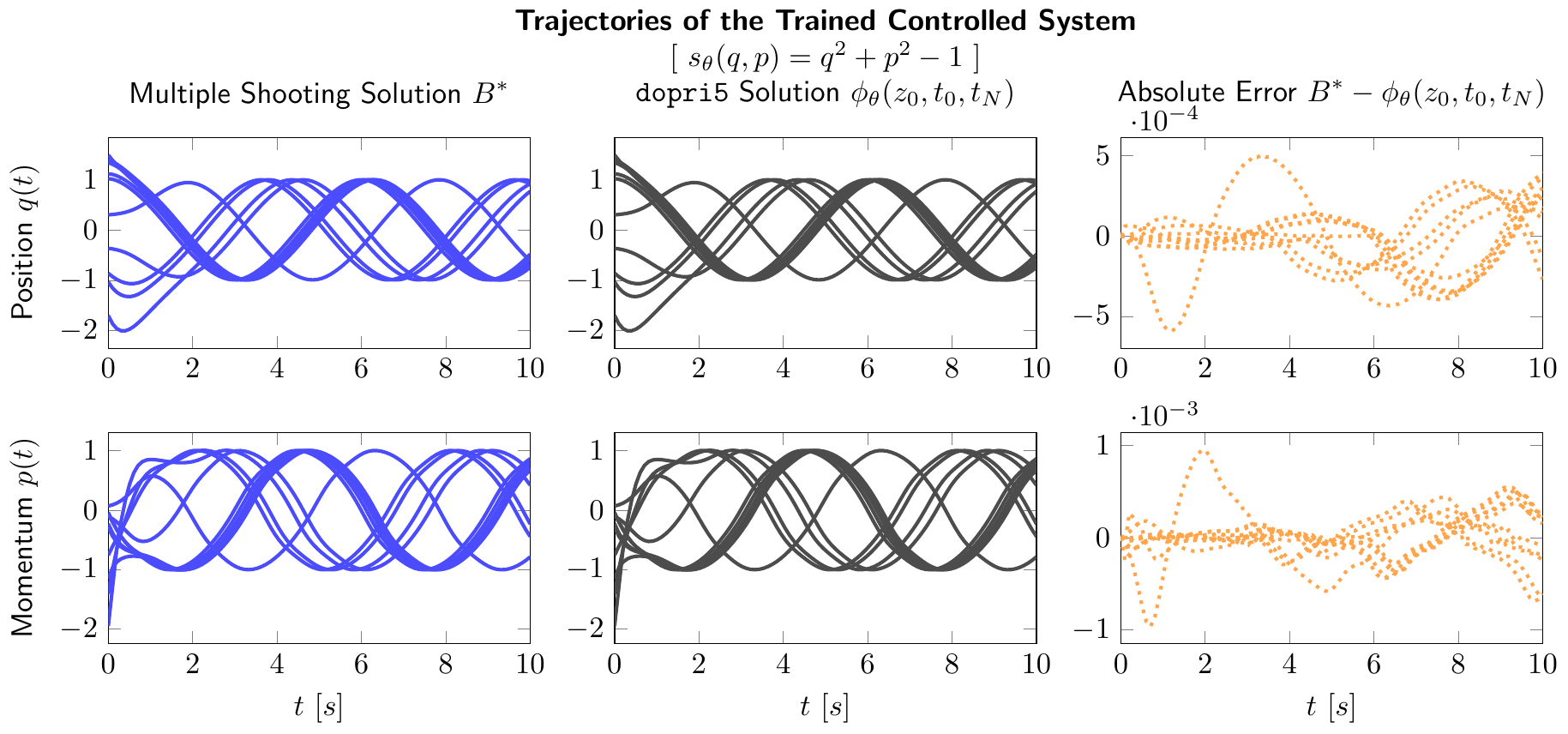}
    \vspace{-8mm}
    \caption{Trained MSL controller on the \textit{circle} experiment. Comparison of the closed--loop trajectories obtained with MSL $B^*$ and the {\tt dopri5} counterpart.}
    \label{fig:cmsl_circle}
\end{figure}

\begin{figure}
    \centering
    \includegraphics[width=\linewidth]{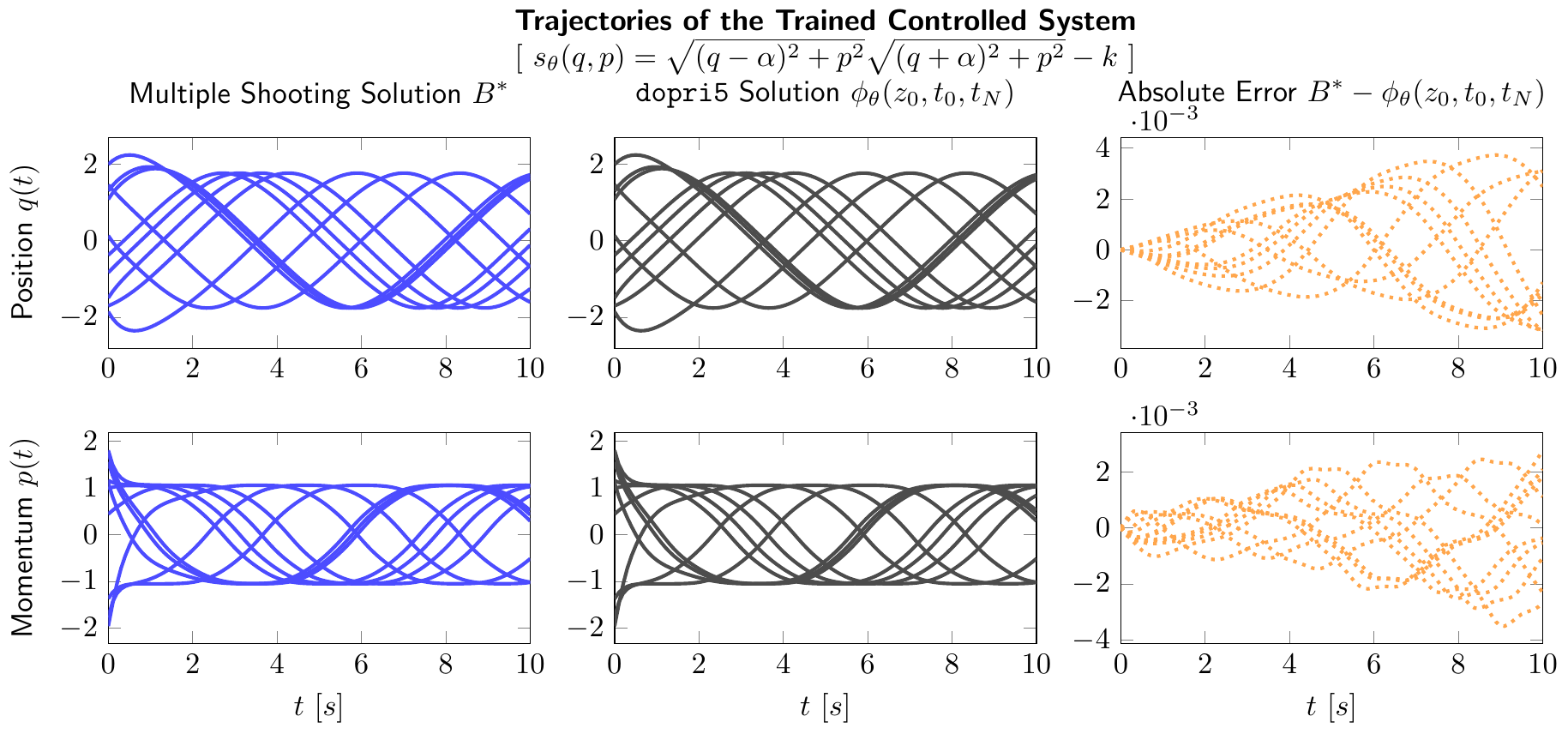}
    \vspace{-8mm}
    \caption{Trained MSL controller on the \textit{circus} experiment. Comparison of the closed--loop trajectories obtained with MSL $B^*$ and the {\tt dopri5} counterpart.}
    \label{fig:cmsl_circus}
\end{figure}

\paragraph{Analysis of results} Figures~\ref{fig:cmsl_circle} and~\ref{fig:cmsl_circus} display the resulting trajectories of the trained MSL in the \textit{circle} and \textit{circus} control tasks. In particular, we compared the last MSL forward solution $B^*$ with the trajectories obtained with the accurate sequential solver with the trained $\pi_\theta$. 

\begin{figure}
    \centering
    \includegraphics[width=.475\linewidth]{figures/cmsl_stream.pdf}
    \hspace{4mm}
    \includegraphics[width=.475\linewidth]{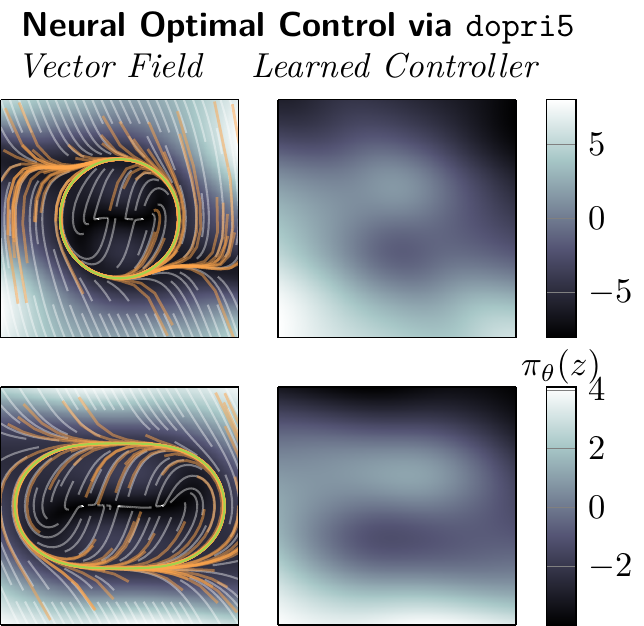}
    \vspace{1mm}
    \caption{Comparison between the learned controllers and closed loop vector fields for the MSL and Neural ODE baseline, in different tasks.}
    \label{fig:dopri_vs_cmsl}
\end{figure}

We also notice that the MSL and Neural ODE baseline converge to very similar controllers and closed--loop vector fields, as it is shown in Fig.~\ref{fig:dopri_vs_cmsl}.

\begin{figure}
    \centering
    \includegraphics{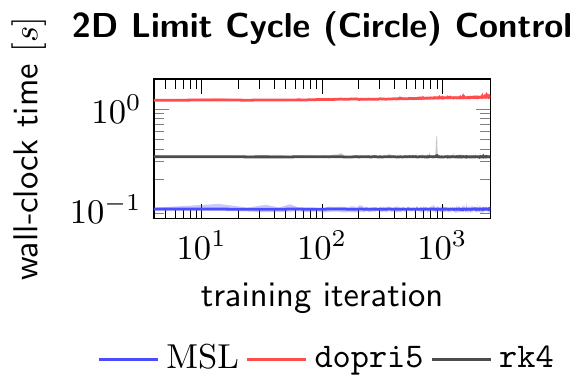}
    \caption{Wall-clock time of complete training iteration (forward/backward passes + GD update) for different solvers on the \textit{circle experimen}}
    \label{fig:wall_clock_control2}
\end{figure}

In Fig.~\ref{fig:wall_clock_control2}, we report the wall-clock times of each forward--backward passes across training iterations. It can be noticed how MSLs encompass sequential approaches with a 10x speedup compared to {\tt dopri5} (even though maintaining a similar accuracy in the solutions) and a 3x speedup w.r.t. the sequential {\tt rk4} solver with the same number of steps per sub--interval.
\subsection{Neural Optimal Boudary Control of the Timoshenko Beam}\label{asec:exp_pde}
With this experiment we aimed at showing the scaling of {\color{blue!70}{\tt fw sensitivity} MSL} to higher--dimensional regimes in a neural--network optimal control tasks. In particular, we wished to investigate if the acceleration property of \textit{one--step} MSLs established by Theorem \ref{th:track} holds when the system state has hundreds of dimensions.

\paragraph{Model and training} We kept an identical training setup to the limit cycle control task of \ref{asec:exp_ctrl}. However, we chose a time horizon of $5s$ and we fixed $N=500$ shooting parameters. We only compared the proposed MSL model to the sequential {\tt rk4} as we empirically noticed how {\tt dopri5} was extremely slow to perform a single integration of the discretized PDE (possibly due to the stiffness of the problem) and was also highly numerically unstable (high rate of \textit{underflows}). 

We implemented a software routine based on the {\tt fenics} \cite{alnaes2015fenics} computational platform to obtain the finite--elements discretization (namely, matrices $A$ and $B$ in \eqref{aeq:fe_model}) of the PDE given the physical parameters of the model, the number of elements, and the initial condition of the beam. We chose a 50 elements discretization of the Timoshenko PDE for a total of 200 dimensions of the discretized state $\ubar z(t)$ and we initialized the distributed state as $z(x, 0) = [\sin(\pi x), \sin(3\pi x), 0, 0]$.

\begin{figure}[h]
    \centering
    \includegraphics{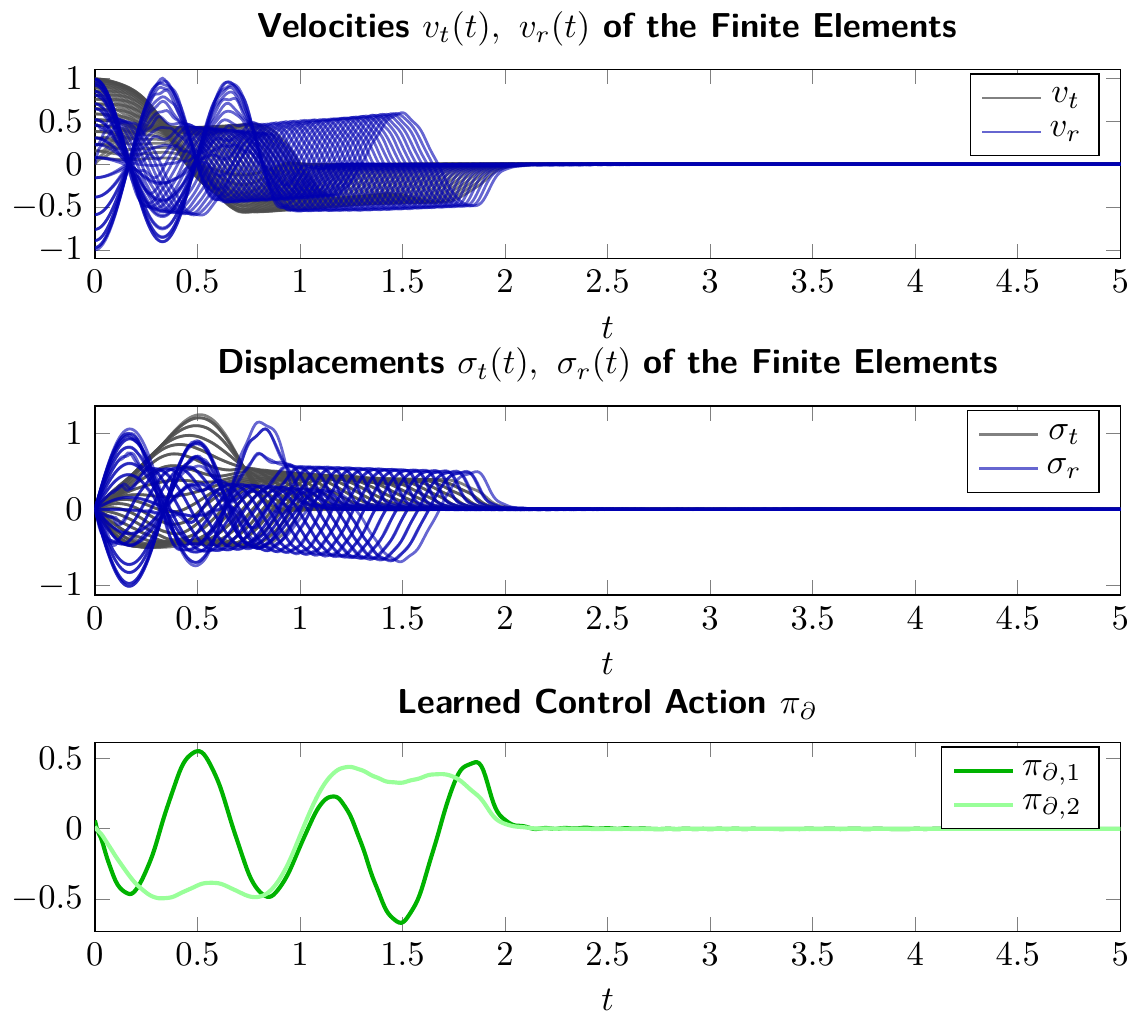}
    \caption{Trajectories of the finite elements states and learned control policy along the trajectory.}
    \label{fig:pde_traj}
\end{figure}

\begin{figure}[h]
    \centering
    \includegraphics{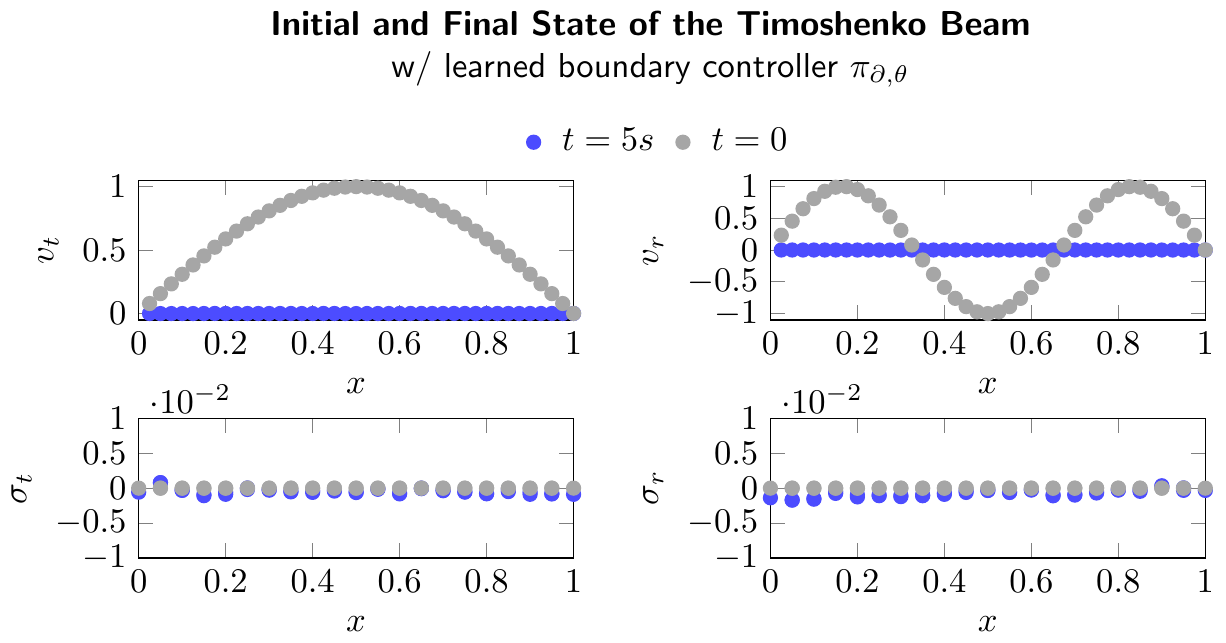}
    \caption{Initial and final (discretized) state of the controlled Timoshenko beam after training with MSL.}
    \label{fig:pde_if}
\end{figure}

Since the experiment focus was the numerical performance of MSL training compared to Neural ODE baselines, we considered a simple stabilization task where the cantilever beam had to be straight. For this reason we selected the following loss criterium
\[
    L_\theta = \frac{1}{N}\sum_{n=0}^{N}\left(\|b^*_{n,\sigma_r}\|_2 + \|b^*_{n,\sigma_t}\|_2\right) + \alpha\left\|\pi_{\partial,\theta}(b^*_{n})\right\|_1
\]
being $b^*_{n,\sigma_r}$, $b^*_{n,\sigma_t}$ the portions of the shooting parameters corresponding to $\ubar \sigma_r$ and $\ubar\sigma_t$, respectively. The boundary controller was designed as a four-layers neural network with 16 neurons per layer, {\tt softplus} activation on the first two hidden layers and hyperbolic tangent activation on the third.
\paragraph{Analysis of results}
{We report additional experimental results. Figure~\ref{fig:pde_traj} displays the trajectories of the system with the learned boundary control policy. It can be seen how the displacements variables for each of the finite elements swiftly goes to zero (straight beam configuration) with zero velocity proving the effectiveness of the proposed model. Finally, Fig.~\ref{fig:pde_if}} shows the initial and final configurations of the finite elements over the spatial domain $x\in[0, 1]$.
\subsection{Fast Neural CDEs for Time Series Classification}\label{asec:exp_cde}
\paragraph{Dataset}
We consider sepsis prediction with data from the {\tt PhysioNet} 2019 challenge. In particular, the chosen dataset features $40335$ variable length time series of patient features. The task involves predicting whether patients develop sepsis over the course of their \textit{intensive care unit} (ICU) stay, using the first 72 hours of observations. Since positive and negative classes are highly imbalanced, we report \textit{area under the receiver operating characteristic} ({\tt AUROC}) as task performance metric. For more details see \citep{kidger2020neural}, which contains the experimental setup followed in this work, and \citep{clifford2015physionet} for more details on the dataset and task.
The data split is performed according to \citep{kidger2020neural} with $70\%$ train, $15\%$ validation and $15\%$ test. 
The $70\%$ split corresponds to $28233$ time series, which in this experiment is taken as batch size to enable application of tracking MSLs relying on Theorem \ref{th:track}.
\paragraph{Models and training}
All model hyperparameters are collected from \citep{kidger2020neural} for a fair comparison. We train a standard \textit{neural controlled differential equation} (Neural CDE) and an equivalent Neural CDE solved with a {\color{orange!70} zeroth--order MSL}. Both baseline and MSL Neural CDEs use standard \textit{reverse mode autodiff} to compute gradients.
We train for $1000$ epochs (here equivalent to iterations due to full--batch training) with a learning rate of $10^{-4}$ for {\tt AdamW} \citep{loshchilov2017decoupled} and with weight decay regularization of $0.03$.